\documentclass[conference]{IEEEtran}
\usepackage{times}

\usepackage[numbers]{natbib}
\usepackage{multicol}
\usepackage[bookmarks=true]{hyperref}
\usepackage{hyperref}
\usepackage{url}
\usepackage{booktabs}
\usepackage{amssymb}
\usepackage{pifont} %
\usepackage{graphicx}
\usepackage{subcaption}
\usepackage{xcolor}
\usepackage{fontawesome}
\usepackage{amsmath, algorithm, algpseudocode}
\usepackage{adjustbox}
\usepackage{wrapfig} %
\usepackage{colortbl} %
\usepackage{array}    %
\begin{document}

\title{Demonstrating GPU Parallelized Robot Simulation and Rendering for Generalizable Embodied AI with ManiSkill3}

\author{Stone Tao$^{1,3}$, Fanbo Xiang$^{1,3}$, Arth Shukla$^{1,3}$, Yuzhe Qin$^1$, Xander Hinrichsen$^1$, 
Xiaodi Yuan$^{1,3}$, \\ Chen Bao$^2$, Xinsong Lin$^1$, Yulin Liu $^{1,3}$, Tse-kai Chan$^1$, Yuan Gao$^1$, Xuanlin Li$^1$, Tongzhou Mu$^1$,\\ Nan Xiao$^1$, Arnav Gurha$^1$, Viswesh Nagaswamy Rajesh$^1$, Yong Woo Choi$^1$, Yen-Ru Chen$^1$,\\ Zhiao Huang$^{1,3}$, Roberto Calandra$^4$, Rui Chen$^5$, Shan Luo$^6$, Hao Su$^{1,3}$\\
$^{1}$University of California San Diego, $^{2}$Carnegie Mellon University, $^{3}$Hillbot, $^{4}$TU Dresden\\ $^{5}$Tsinghua University, $^{6}$King's College London\\
}

\maketitle
\begin{figure*}[t!]
    \centering
    \makebox[\textwidth]{\includegraphics[width=1\linewidth]{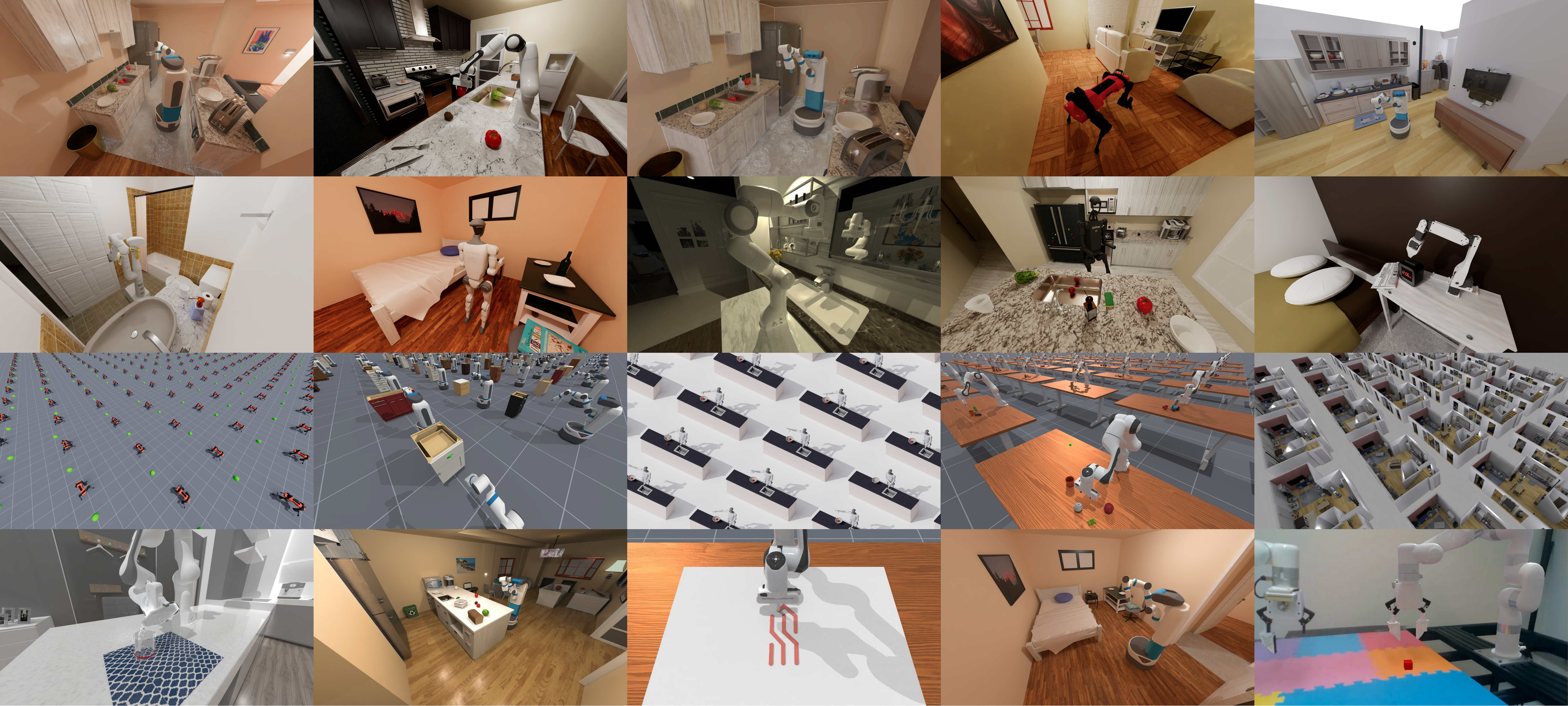}}
    \caption{Multiple distinct task categories are displayed, ranging from room-scale tasks to humanoid interactions and drawing tasks. Majority of tasks shown are GPU-parallelized, simulating + rendering at state-of-the-art speeds and GPU memory efficiency. Scenes are from ReplicaCAD and AI2-THOR.}
    \label{fig:task_categories}
\end{figure*}
\begin{abstract}
Simulation has enabled unprecedented compute-scalable approaches to robot learning. However, many existing simulation frameworks typically support a narrow range of scenes/tasks and lack features critical for scaling generalizable robotics and sim2real. We introduce and open source ManiSkill3, the fastest state-visual GPU parallelized robotics simulator with contact-rich physics targeting generalizable manipulation. ManiSkill3 supports GPU parallelization of many aspects including simulation+rendering, heterogeneous simulation, pointclouds/voxels visual input, and more. GPU Simulation with rendering on ManiSkill3 uses 2-3x less GPU memory usage than other platforms and achieves up to 30,000+ FPS in benchmarked environments due to minimal python/pytorch overhead in the system, simulation on the GPU, and the use of the SAPIEN parallel rendering system. Tasks that used to take hours to train can now take minutes. We further provide the most comprehensive range of GPU parallelized environments/tasks spanning 12 distinct domains including but not limited to mobile manipulation, drawing, humanoids, and dextrous manipulation in realistic scenes designed by artists or real-world digital twins. In addition, millions of demonstration frames are provided from motion planning, RL, and teleoperation. ManiSkill3 also provides a comprehensive set of baselines that span popular RL and learning-from-demonstrations algorithms. 
\end{abstract}

\IEEEpeerreviewmaketitle

\section{Introduction}

One of the grand challenges of robotics is robust and generalized manipulation. However, unlike vision and language research, there are still no good datasets for robotic manipulation that can be trained on. One approach has been to create human-scalable real-world teleoperation tools \citep{DBLP:journals/corr/abs-2406-10454-human-plus, cheng2024tv} to then perform imitation learning. Another is to set up real-world reinforcement learning to fine-tune offline trained policies \citep{feng2023finetuning}.
However, real-world imitation learning approaches require enormous amounts of data that are infeasible to collect efficiently at low costs only to achieve relatively low success rates that are otherwise impractical for real-world deployment \citep{zhao2024aloha}. Real-world reinforcement learning approaches are promising, but require extensive setup in the real world to generate real-world rewards/success and environment resets.

GPU parallelized simulations such as Isaac \citep{DBLP:conf/nips/MakoviychukWGLS21-isaac} and Mujoco's MJX \citep{todorov2012mujoco} have made massive advancements in solving some robotics problems such as robot locomotion by training in large-scale GPU parallelized simulations with reinforcement learning (RL) \citep{DBLP:conf/corl/RudinHR021}. GPU parallelized simulation makes data incredibly cheap to generate. However, when it comes to manipulation, success is often limited to narrower ranges of manipulation tasks and typically requires strong state estimation \citep{DBLP:conf/icra/HandaAMPSLMWZSN23-dextreme} to replace visual inputs like RGB or pointcloud. Existing GPU simulators have limitations that hinder the generalization and scalability of previous work. These simulators lack support for heterogeneous simulation, where each parallel environment contains different scenes. Additionally, they often don't support fast parallel rendering capabilities. As a result, algorithms like reinforcement learning (RL) that operate on visual input train too slowly to be practical. We propose ManiSkill3 to address past limitations and open source the framework under the Apache-2.0 license, building upon past work in ManiSkill 1 and 2 \citep{mu2021maniskill, gu2023maniskill2}. The majority of features presented in this paper are already open-sourced. Website: \href{https://maniskill.ai/}{maniskill.ai/}. Documentation/Tutorials: \href{https://maniskill.readthedocs.io/en/latest/user_guide/index.html}{maniskill.readthedocs.io/}. Video demo gallery: \href{https://maniskill.readthedocs.io/en/latest/user_guide/demos/gallery.html}{maniskill.readthedocs.io/en/latest/user\_guide/demos/gallery.html}.

The core contributions of ManiSkill3 that set it apart from existing simulators are as follows:

\textbf{1) State-of-the-art GPU Parallelized Simulation and Rendering}: RL algorithms like PPO \citep{DBLP:journals/corr/SchulmanWDRK17-ppo} can now solve visual tasks faster than it would have taken on other simulators due to fast parallel rendering and low overhead in the system design of ManiSkill3, leading to highly efficient use of the GPU. Depending on task the simulation + rendering FPS can reach up to 30,000+, massively accelerating visual data collection relative to most other simulators. Importantly ManiSkill3 maintains extremely low GPU memory usage, typically 2-3x lower than that of other simulators which enables on device visual RL and larger neural networks during training.

\textbf{2) Most comprehensive range of environments with 12 different categories of environments and 20+ different robots provided out of the box}: ManiSkill3 out of the box provides a diverse set of different types of environments including but not limited to mobile manipulation, room-scale scenes, drawing, and humanoid/bi-manual manipulation. We further support 20+ different robot embodiments out of the box such as quadrupeds, floating grippers, humanoids, and dextrous hands. Furthermore we support several sim2real and real2sim setups for manipulation. Importantly, extensive documentation/tutorials are provided to teach users on how to add new environments/robots, as well as how to make open-source contributions to expand the repository of simulated tasks/robots. A core focus of ManiSkill3 is not on building a lot of environments in each category, but instead building many template/examples that users can then build on top of themselves for their own use-cases.

\textbf{3) Heterogeneous Simulation for Generalizable Learning}: ManiSkill3 makes it possible to simulate and render completely different objects, articulations, even entire room-scale scenes in each parallel environment. This is done thanks to a data-oriented system design and easy-to-use API to manage GPU memory of objects/articulations even if they may have different degrees of freedom.

\textbf{4) Simple Unified API to Easily Manage and Build GPU Simulated Tasks}: ManiSkill3 distinguishes itself from other GPU-parallelized robotics simulators and benchmarks by offering a user-friendly API for creating diverse robotics environments. Improvements include object-oriented APIs and the elimination of complex tensor indexing. The platform provides feature-rich tooling to streamline various operations, such as domain randomization (e.g., camera poses, object materials), trajectory replay, controller action conversion, and more.

\textbf{5) Scalable Dataset Generation Pipeline from Few Demonstrations}: For tasks in ManiSkill3 where reward design is difficult, we provide a pipeline that leverages demonstration efficient, wall-time fast, online imitation learning algorithms, to learn a generalized neural network policy from a few teleoperated/hardcoded demonstrations. The generalized task-specific neural network policy is then used to rollout many more demonstrations to form larger datasets.

\section{Related Work}
\definecolor{darkgreen}{RGB}{0,150,0} %

\newcommand{\warn}{\textcolor{orange}{\faExclamationTriangle}} %

\newcommand{\cmark}{\textcolor{darkgreen}{\ding{51}}} %
\newcommand{\xmark}{\textcolor{red}{\ding{55}}}   %

\begin{table*}[!htbp]
\centering
\begin{tabular}{p{1.8in}ccccccc}
\hline
Feature & \rotatebox{45}{ManiSkill3} & \rotatebox{45}{Isaac Lab} & \rotatebox{45}{RoboCasa} & \rotatebox{45}{RLBench} & \rotatebox{45}{OmniGibson} & \rotatebox{45}{Habitat} & \rotatebox{45}{AI2THOR} \\
\hline
Parallelized Simulation         &  \cmark & \cmark & \xmark  &  \xmark & \xmark & \xmark & \xmark \\
Parallelized Rendering & \cmark & \cmark  &  \xmark &  \xmark & \xmark & \xmark & \xmark\\
Parallelized Heterogeneous Simulation & \cmark & \xmark & \xmark & \xmark & \xmark & \xmark & \xmark \\
Large Scale Demonstrations & \cmark & \xmark & \cmark & \cmark  & \xmark & \xmark & \xmark \\
Realistic Object Physics & \cmark & \cmark & \cmark  & \cmark  & \cmark  & \xmark & \xmark  \\
Photorealistic Rendering & \cmark & \cmark & \cmark  &  \xmark &  \cmark & \xmark & \xmark \\
Room-Scale Scenes & \cmark & \cmark & \cmark & \xmark & \cmark  & \cmark & \cmark  \\
Visual RL Baselines & \cmark & \cmark & \xmark & \xmark & \xmark & \cmark & \cmark \\
Vision-based sim2real setups & \cmark & \xmark & \xmark & \xmark & \xmark & \xmark & \xmark \\
Trajectory replay/conversion & \cmark & \xmark & \xmark & \xmark & \xmark & \xmark & \xmark \\
Task Categories & 12 & 5 & 2 & 1 & 2 & 1 & 1\\
Interactive GUI & \cmark & \cmark & \xmark & \xmark & \xmark & \cmark & \xmark \\
\hline
\end{tabular}
\caption{Comparison of major features across different open-source robotics frameworks/tools.}
\label{tab:sim_comparison}
\end{table*}
\textbf{Robotics Simulation Frameworks:} Isaac Lab (previously called Isaac Orbit) \citep{mittal2023orbit} and Mujoco \citep{todorov2012mujoco} are some open-source general-purpose rigid-body GPU parallelized robotics simulators. Isaac Lab and Brax \citep{brax2021github} (which supports the Mujoco MJX backend) are the most similar to ManiSkill3 in that they provides out of the box environments for reinforcement learning/imitation learning, as well as APIs to build environments. There are robotics frameworks like Robocasa \citep{robocasa2024}, Habitat, \citep{szot2021habitat}, AI2THOR \citep{ai2thor}, OmniGibson \citep{li2022behavior}, RLBench \citep{james2019rlbench} that only have CPU sim backends and thus run magnitudes slower than Isaac Lab/ManiSkill3, limiting researchers to often only explore imitation learning/motion planning approaches instead of reinforcement learning/online learning from demonstrations methods. Isaac Lab relies on the closed source Isaac Sim framework for GPU parallelized simulation and rendering whereas ManiSkill3 relies on the open-source SAPIEN \citep{Xiang_2020_SAPIEN} for the same features. Brax/Mujoco uses the MJX backend and currently does not have parallel rendering. Both Isaac Lab and ManiSkill3 use PhysX for GPU simulation. Note that the environments and baselines leveraging the fast parallel rendering in ManiSkill3 is concurrent work to Isaac Lab.

\textbf{Robotics Datasets:} Amongst existing datasets there are typically two kinds, real-world and simulated datasets. Open-X \citep{open_x_embodiment_rt_x_2023} is one of the largest real-world robotics datasets but suffers from issues with inconsistent data labels and overall poor data quality. DROID \citep{khazatsky2024droid} addresses some of Open-X's problems by using a consistant data collection platform. However, both Open-X and DROID require immense amounts of human labor to collect data and are inherently difficult to scale up to the sizes of typical vision/language datasets. Among simulated datasets, frameworks like AI2-THOR \citep{ai2thor}, and OmniGibson \citep{li2022behavior} have complex room-scale scenes but do not readily provide demonstrations or ways to generate large-scale demonstrations for use in robot learning. Robocasa has a myriad of tasks and realistic room-scale scenes, but further leverages MimicGen \citep{mandlekar2023mimicgen} to scale human teleoperated demonstrations by generating new demonstrations. 

ManiSkill3 sources large-scale demonstrations through a combination of different methods. For easier tasks, motion planning and rewards for RL are used to generate demonstrations. For more complex tasks without easily defined motion planning scripts or reward functions, ManiSkill3 relies on online learning from demonstrations algorithms like RLPD \citep{DBLP:conf/icml/BallSKL23-rlpd} and RFCL \citep{DBLP:conf/iclr/TaoSC024-rfcl}, which are more flexible compared to MimicGen used in Robocasa as MimicGen makes a number of assumptions about the task (end-effector action spaces, little to no geometric variations, engineered task stage indicators).

\section{Core Features of ManiSkill3}

ManiSkill3 is the most feature-rich GPU simulation framework compared to popular alternatives as shown in Table \ref{tab:sim_comparison}. For the largest features, we detail them in the subsections below.

\subsection{Diverse Tasks Supported Out Of The Box}

The design of ManiSkill3 enables support for many different kinds of task categories via a flexible task-building API. Of the existing popular robotics simulators ManiSkill3 supports the most categories of different tasks. Concretely we categorize the 12 distinct categories as follows: Table top manipulation, mobile manipulation, room-scale scenes for manipulation, quadruped/humanoid locomotion, humanoid/bi-manual manipulation, multi-agent robotics, drawing/cleaning, dextrous manipulation, vision-tactile manipulation, classic control, digital twins, and soft body manipulation environments. The majority of these tasks are GPU parallelized and can be rendered fast in parallel as well, with examples of the tasks shown in Fig. \ref{fig:task_categories}. Each of these task categories have various optimizations done to run more accurately and/or faster. Other simulators typically support a smaller subset of the type of tasks ManiSkill3 supports easily. Additional details on the exact optimizations/implementations and available robots are detailed in Appendix \ref{appendix:tasks}.
\begin{figure*}[ht]
    \centering
    \includegraphics[width=\linewidth]{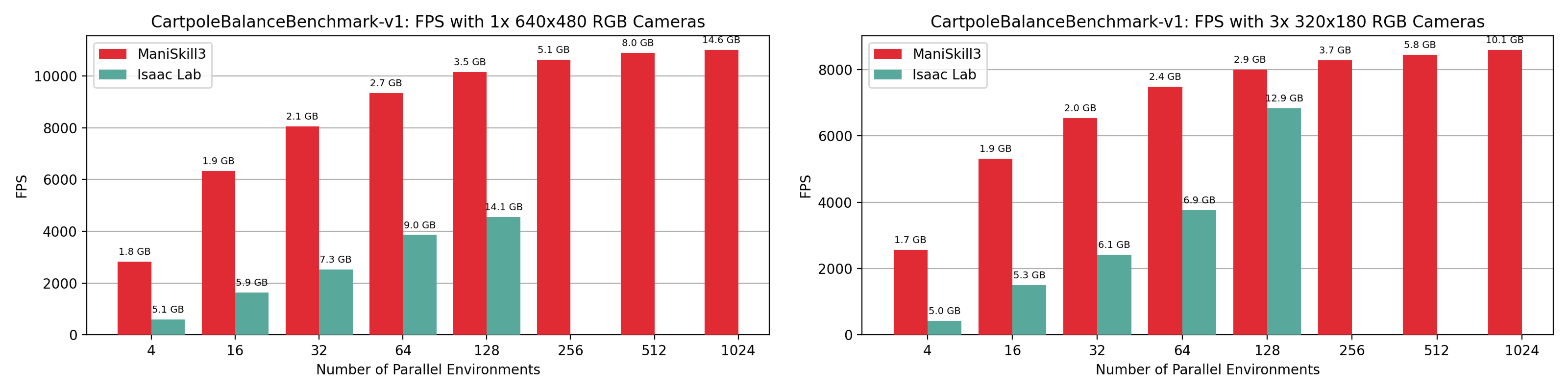}
    \caption{GPU Simulation+Rendering of RGB speeds of the Cartpole environment with different camera setups ManiSkill3 and Isaac Lab. Annotated numbers indicate GPU memory usage, with no data points beyond 128 environments for Isaac Lab due to running out of GPU memory. Note that this rendering setting mimics that of real world datasets collected in Open-X and Droid. Speed is dependent on a few factors, primarily the number of objects, geometry complexity of each object, as well as simulation/rendering configurations which can be tuned for speed or accuracy. As a result, it is possible the numbers/trends here may not hold for every environment.}
    \label{fig:parallel_sim_render_perf_vs_isaac}
\end{figure*}

\begin{figure}[h]
    \centering
    \includegraphics[width=0.75\linewidth]{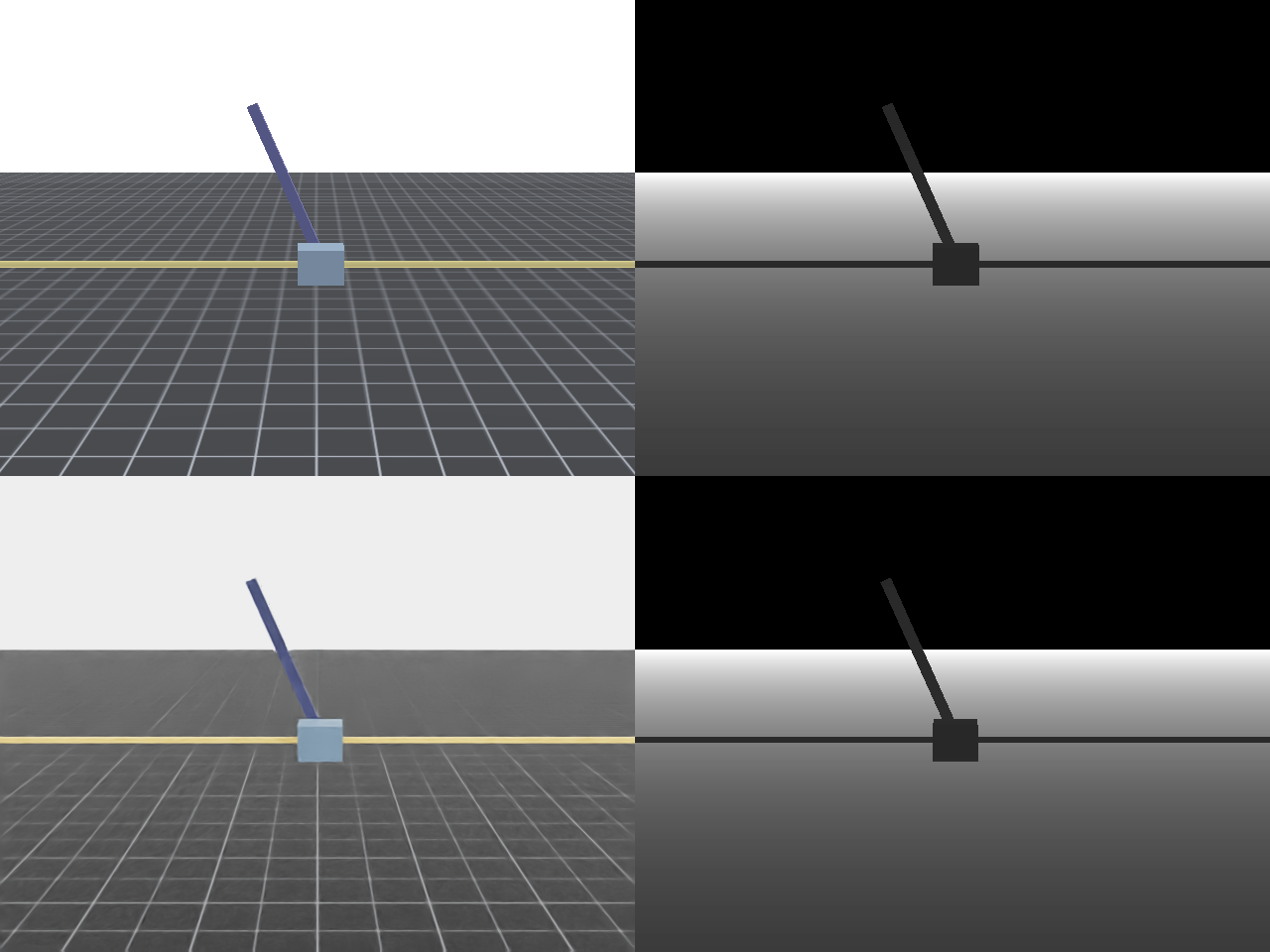}
    \caption{Comparison of ManiSkill3 (Top row) and Isaac Lab (Bottom row) parallel rendering 640x480 RGB and depth image outputs of the Cartpole benchmark task.}
    \label{fig:cartpole-visual-comparison}
\end{figure}
\subsection{GPU Parallelized Simulation and Rendering}
\label{sec:sim+rendering}
\begin{figure}[h]
    \centering
    \includegraphics[width=\linewidth]{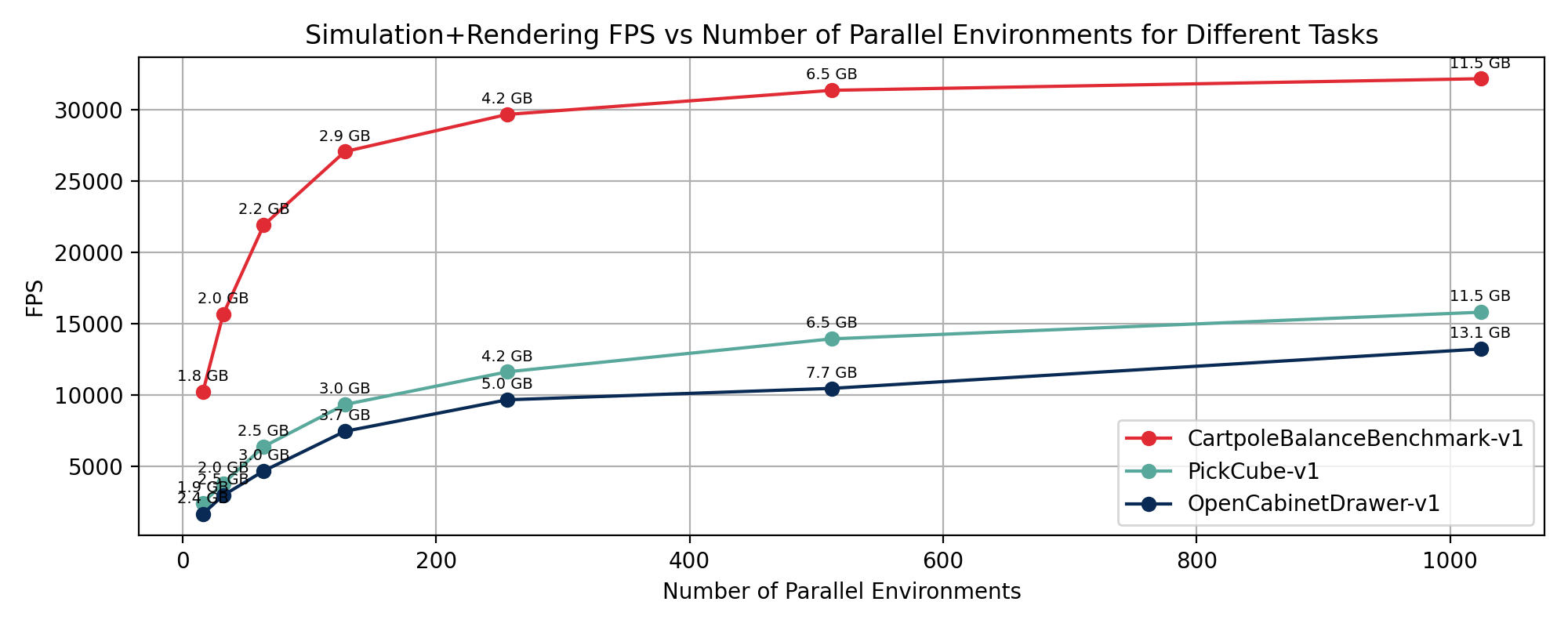}
    \caption{GPU Simulation+Rendering speeds of various tasks with a single 128x128 resolution camera with a simulation frequency of 120 and control frequency of 60, meaning the camera renders ever 2 sim steps. RGB, depth, and segmentation data are all simultaneously being rendered. The only big variations between environments of the three curves are the objects and robots being simulated. }
    \label{fig:parallel_sim_render_perf}
\end{figure}
ManiSkill3 distinguishes itself from its predecessors and other robotics simulators by offering robust support for GPU-parallelized simulation and rendering. ManiSkill3 is the first general benchmark to enable fast RL from visual inputs on complex robot manipulation tasks, with Isaac Lab recently adding a similar feature. Tasks such as picking up a cube or controlling a quadruped to reach a goal from pixel inputs are now solved on the order of minutes instead of hours. RL training results/speed are detailed in Section \ref{sec:rl_results}.

The performance results shown in Figure \ref{fig:parallel_sim_render_perf} are the results after simulating + rendering RGB, depth, and segmentation data simultaneously for various tasks. In terms of speed and GPU memory use, Figure \ref{fig:parallel_sim_render_perf_vs_isaac} shows ManiSkill3 outperforms Isaac Lab, particularly when it comes to rendering common real-world camera resolutions which can be important for sim2real and real2sim transfer. In particular, with 128 parallel environments for the benchmarked task, ManiSkill3 uses just 3.5GB of GPU memory whereas Isaac Lab uses 14.1GB. 
The memory efficiency of the ManiSkill3 platform allows for more room for e.g. RL replay buffers or larger neural network models such as large vision language action models. Training and inference can be kept extremely optimized on a single GPU as a result without needing to store any data on the CPU. From experimentation with visual RL, we find that GPU memory efficiency becomes much more important as the FPS gains from more parallel environments become marginal. GPU memory efficiency is especially important for off-policy algorithms like TD-MPC2 \citep{hansen2024tdmpc2} and SAC \citep{DBLP:conf/icml/HaarnojaZAL18-sac} that typically maintain replay buffer sizes on the order of $10^5 \sim 10^6$ frames. For example storing RGB data from one 128x128 camera would require at least 9GB of GPU memory for a replay buffer of size $200,000$, which can easily lead to out of memory issues. For more in-depth performance benchmarking results and comparisons of rendered outputs, see Appendix \ref{appendix:sim-render-benchmark}. 

We acknowledge that this comparison is not strictly apples-to-apples due to differences in rendering techniques. Isaac Lab employs ray-tracing for parallel rendering, while the ManiSkill3 results are generated using SAPIEN's rasterization renderer (see Figure \ref{fig:cartpole-visual-comparison} for a visual comparison), although ManiSkill3 also supports a ray-tracing mode without parallelization. Ray-tracing generally offers greater flexibility in balancing rendering speed and quality through the adjustment of parameters such as samples per pixel. It's worth noting that the Isaac Lab data presented here uses the fastest rendering settings, although it can be easily tuned to achieve better rendering quality that may be helpful for sim2real. Despite the use of different rendering techniques, we believe this experiment provides a meaningful basis for comparison.

\begin{figure}[h]
    \centering
    \includegraphics[width=\linewidth]{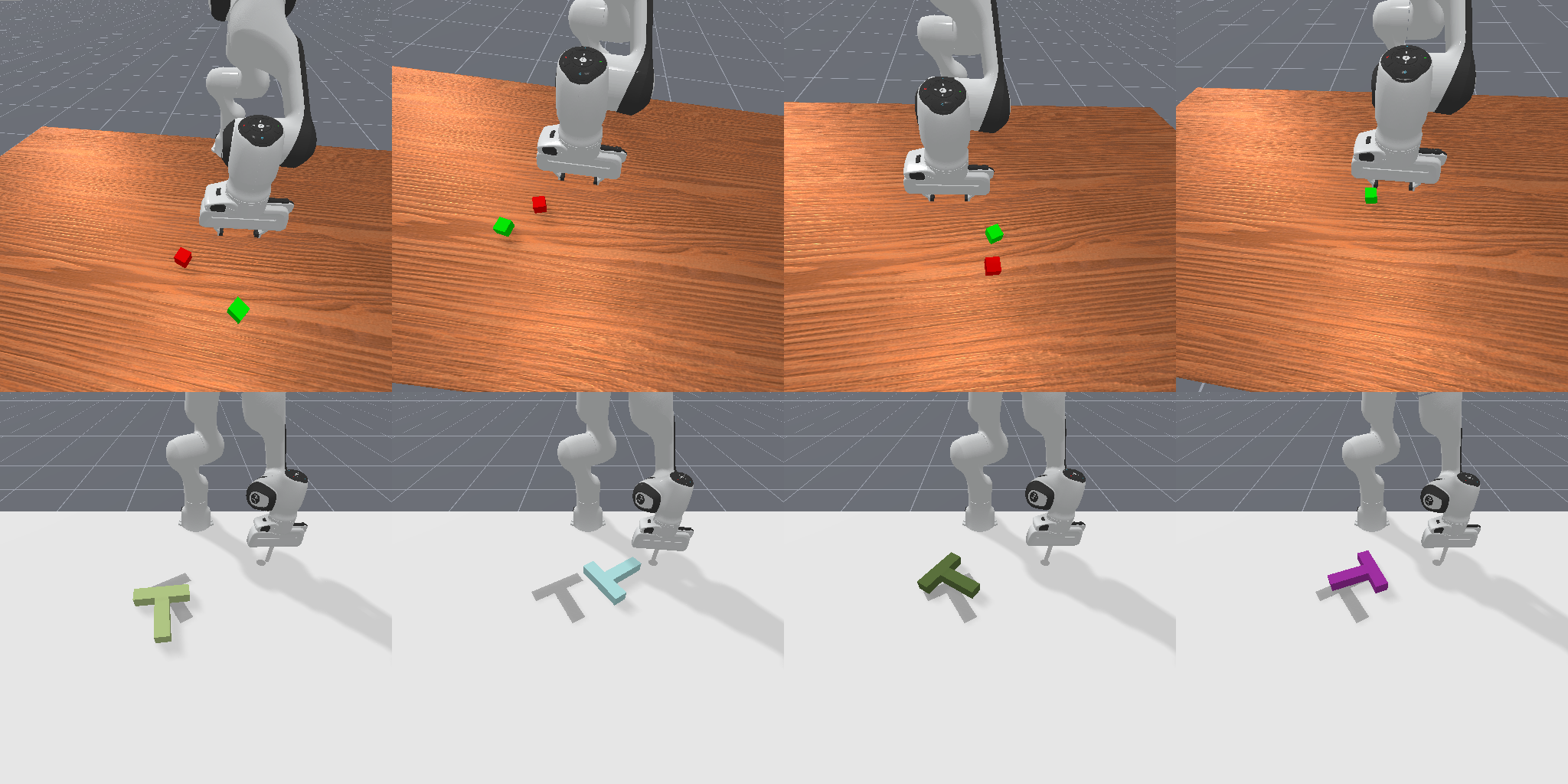}
    \caption{Parallel rendering outputs of 1024 parallel environments for the StackCube and PushT tasks with a subset of 4 them visualized here. Original renders are size 128x128, images shown are up-scaled for clarity. Top-row shows camera pose randomization and bottom row shows texture randomization, renderered depth/segmentation data is not shown here.}
    \label{fig:parallel_domain_randomize}
\end{figure}

GPU parallelized simulation and rendering enable an entirely new regime of running efficient domain randomizations. For example you can now quickly render over a 1000 different cameras, each with different extrinsics/intrinsics, mounted/fixed, as well as randomize object textures in each of the parallel environments. A subset of 4 out of 1024 environments renders are shown in Figure \ref{fig:parallel_domain_randomize} with different settings. This type of visual diversity in simulation enables much faster training of more visually robust policies and is critical for sim2real applications. Furthermore, ManiSkill3 supports parallelized rendering of voxel and pointcloud formats necessary for 3D robot learning approaches \citep{Ze2024DP3, huang2023voxposer, shridhar2022peract}.

Finally, ManiSkill3 enables extremely fast visual digital twins. For example, ManiSkill3 implements 4 of the environments in SIMPLER \citep{li24simpler} which are evaluation digital twins that enable the evaluation of generalist robotic policies trained on real-world data like Octo \citep{octo_2023} and Robotics Transformers (RT) \citep{DBLP:journals/corr/abs-2212-06817-rt1}. ManiSkill3 digital twins can evaluate models like Octo at 60x to 100x the speed of the real world without human supervision, approximately 10x faster than the original digital twin implementations in SIMPLER. The speed increase is due to fast and efficient parallel rendering of large camera resolutions (640x480) and flexible GPU parallelized controllers to match most real-world robot arms/manipulators. More details on sim2real/real2sim support are covered in Section \ref{sec:sim2real}.

\subsection{Heterogeneous GPU Simulation}

\begin{figure}[h]
    \centering
    \includegraphics[width=\linewidth]{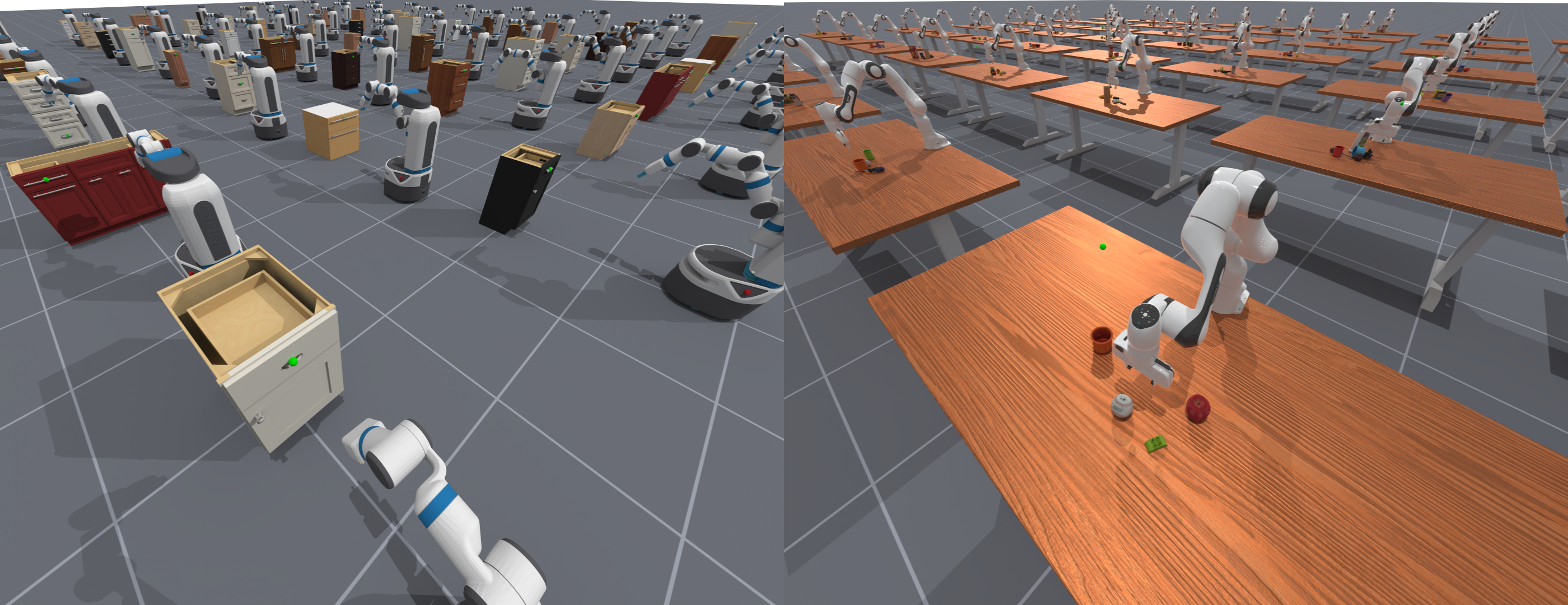}
    \caption{Example tasks in ManiSkill3 showing heterogeneous GPU simulation with different DoF articulations and/or different numbers of objects being simulated in each parallel environment.}
    \label{fig:heterogenous_render}
\end{figure}

ManiSkill3 is so far the only simulation framework that completely supports heterogeneous GPU simulation. This is the feature of being able to simulate different object geometries, different numbers of objects, and different articulations with different DOFs across parallel environments. For example, in the OpenCabinetDrawer task, for each parallel environment, we build a different cabinet (all with different DOFs) and sample a random drawer link that needs to be opened to succeed. In the Pick Clutter YCB task, we sample a different number of YCB objects in each parallel environment and sample one random object out of the clutter as the goal object to pick. ManiSkill3 easily supports this kind of simulation and further supports rendering these different scenes in parallel all at once with an example 3rd view rendering illustrated in Figure \ref{fig:heterogenous_render}. Heterogeneous GPU simulation enables more generalizable learning for manipulation as algorithms like PPO can simultaneously train on every single object from the YCB dataset \citep{DBLP:journals/corr/CalliWSSAD15-ycb} or the PartNetMobility dataset of cabinets \citep{Mo_2019_CVPR}.

\subsection{Teleoperation}

\begin{figure}[htp]
    \centering
    \includegraphics[width=\linewidth]{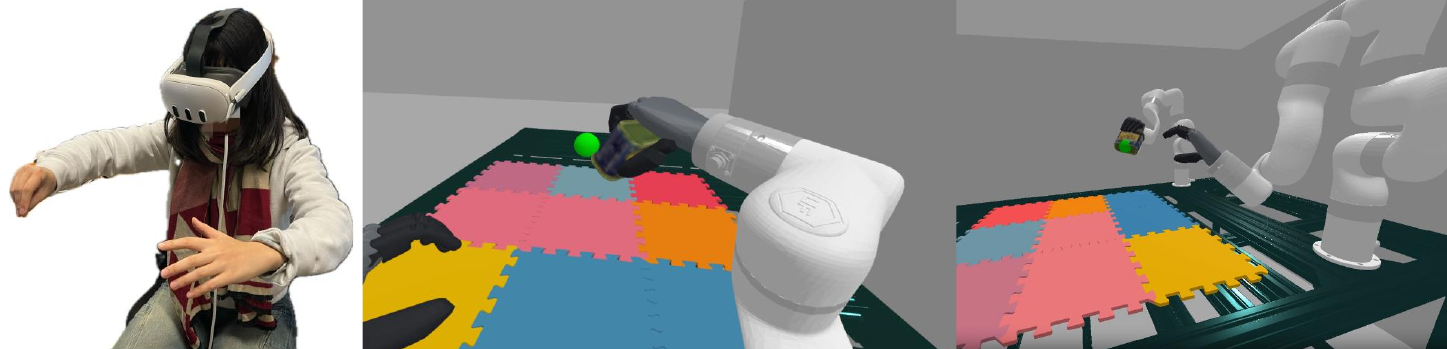}
    \caption{Visualization of VR teleoperation system. Left: A teleoperator using hand poses captured by the Meta Quest 3 headset to control robot motion in real-time. Middle: A 360-degree scene displayed in the VR device, providing immersive sensory feedback. Right: Trajectory replay.}
    \label{fig:vr}
\end{figure}

We provide a Virtual Reality (VR) Teleoperation system that is seamlessly integrated into ManiSkill3 with immersive visual feedback and supports wired connections. ManiSkill3 receives real-time hand pose data from the teleoperator that is translated into corresponding robot actions. At the same time, the system streams 4K stereo video via \href{https://github.com/alvr-org/ALVR}{Air Light VR (ALVR)} to the VR device at 60 Hz, ensuring a smooth and immersive "all-scene view" experience.  This allows the user to explore the entire environment freely, making it possible to teleoperate long-horizon tasks as well as more precise tasks by simply looking closer in the simulation. A crucial control feature is “where your hand is, where the end-effector is”, which tightly aligns the operator’s hand movement with the robot's end-effector. Prior work provide ways to stream video feedback over the internet via a hardware agnostic web app \cite{cheng2024open, ding2024bunny} but tradeoff some ease-of-use for better accessibility (such as long-range remote teleoperation). Our integrated system enables wider viewports and reduce the need for teleoperators to compensate for hand-robot misalignment and visual disparities, visualized in \ref{fig:vr}. Our VR teleop system is also compatible with real-world teleoperation. For real-world operations, we use one or more depth cameras to generate a point cloud of the scene. The point cloud is then rendered and streamed as stereo RGB video to the operator in real time, allowing for effective and immersive control. Example code for configuring this system is provided for teleoperating robot arms or floating robots with multiple grippers/fingers. See Appendix \ref{appendix:teleop} for more details on this teleoperation system.

\subsection{Sim2Real and Real2Sim for Robot Manipulation}
\label{sec:sim2real}

\begin{figure}[h]
    \centering
    \includegraphics[width=\linewidth]{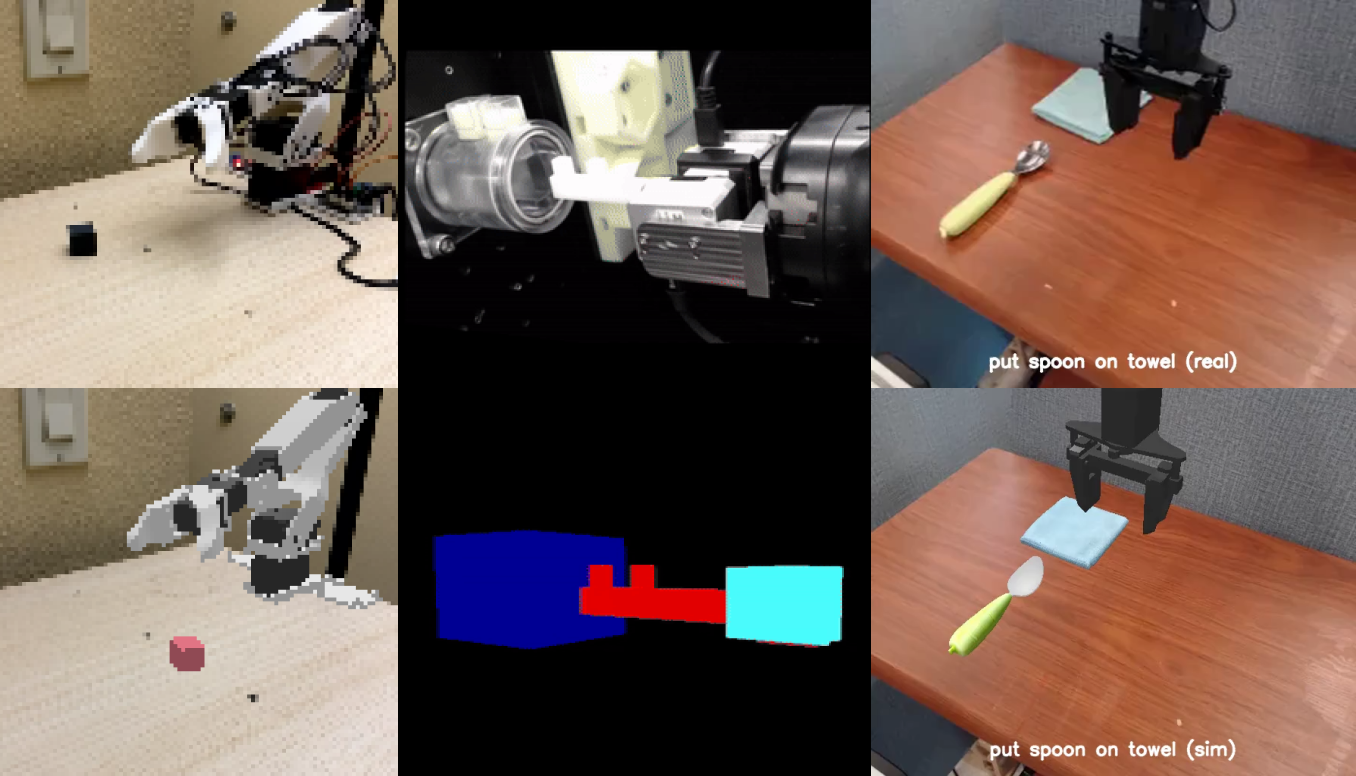}
    \caption{Three different kinds of digital twins in ManiSkill3. Top row shows the real-world setup and bottom row shows the digital twin. Left: Domain randomized digital twin of a cube picking task. Middle: Digital twin of the vision-tactile simulation of a key insertion task. Right: Real2sim digital twin of a spoon placing task.}
    \label{fig:digitaltwins}
\end{figure}

Towards the goal of robust real-world robotics beyond simulation, we verify sim2real and real2sim are both possible using ManiSkill3 via digital twins on some tasks. Figure \ref{fig:digitaltwins} showcases several digital twins supported with real world counterparts. For the cube picking task, we train with RL on simulation images and directly deploy to the real world, achieving a real world success rate of 91.6\% averaged across 3 RL training runs. See Section \ref{sec:sim2realvisionresults} for more details on rigid body manipulation sim2real. For the vision-tactile peg insertion task, we simulate the tactile sensor made of silicone as a softbody and refer readers to the results showcased in the original work by \cite{chen2024tactilesim2real} for those environments, which achieved a 95.08\% success rate in the real world. Finally, for the real2sim digital twins we evaluate Octo and RT-1X on the ManiSkill3 GPU parallelized version of 4 tasks in SIMPLER \citep{li24simpler}. We achieve a correlation between real-world success rates and simulation success rates of 0.9284 and a Mean Maximum Rank Violation (MMRV) value 0.0147, which is close to the original values reported in SIMPLER. See Appendix \ref{appendix:digitaltwins} for more details on real2sim evaluations.

\subsection{Simple Unified API for Building GPU Parallelized Robotics Tasks}
A core reason behind the flexibility of the ManiSkill3 system to support so many different distinct task categories is the clean and simple API for task-building. The simple API also enables users to easily build and customize their own robotics tasks without having to worry about complex GPU memory management details or desigining robot controllers. In addition to the API design \href{https://maniskill.readthedocs.io/en/latest/user_guide/tutorials/custom_tasks/intro.html}{extensive tutorials are provided for customizing tasks}. We describe two general features provided by ManiSkill3 below:

\subsubsection{Object-Oriented API for Articulations, Links, Joints, and Actors}
\begin{figure*}[h]
    \centering
    \includegraphics[width=0.75\linewidth]{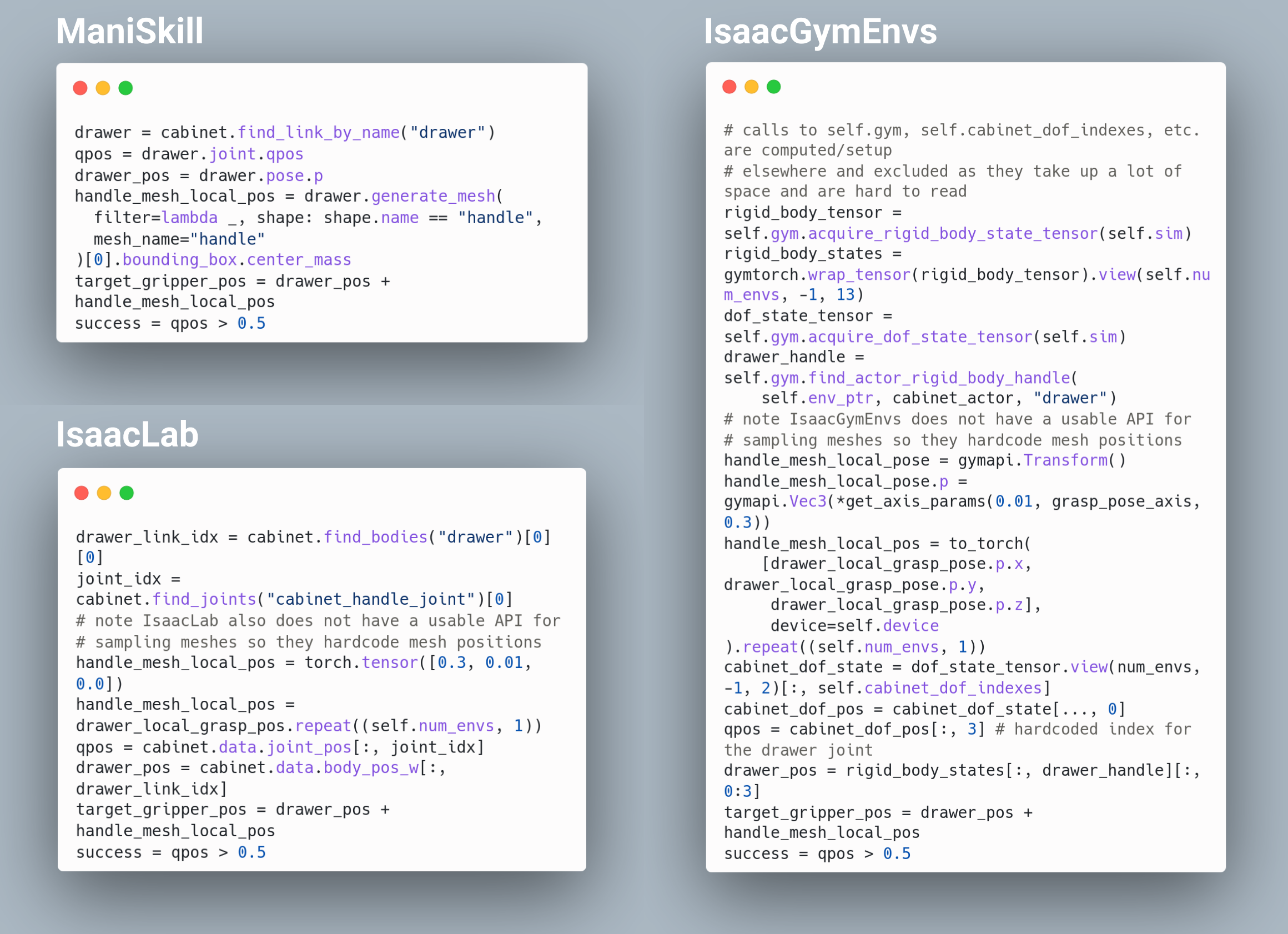}
    \caption{Code comparison for computing a grasp position on a cabinet handle and the joint angle of the cabinet drawer in 3 different GPU simulation frameworks.}
    \label{fig:cabinet-code-comparison}
\end{figure*}
\begin{figure*}[h]
    \centering
    \includegraphics[width=0.82\linewidth]{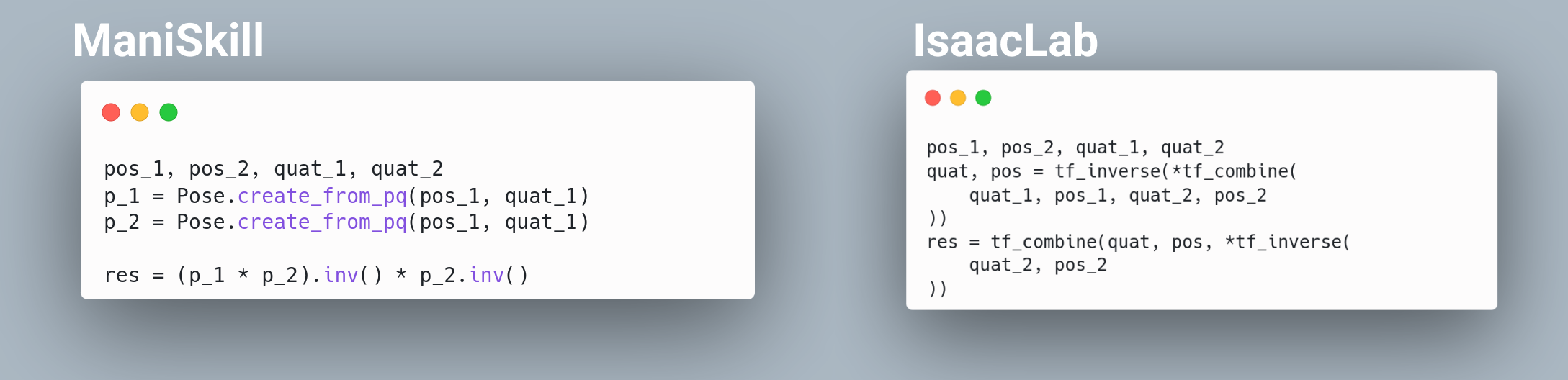}
    \caption{Code comparison for manipulating batched poses}
    \label{fig:pose-code-comparison}
\end{figure*}
ManiSkill3 is the only framework with a complete object-oriented API around the high-level articulations/actors down to individual links/joints and meshes. In contrast, IsaacGymEnvs requires users to instantiates relevant GPU buffers for holding articulation state such as root pose and joint angles. Isaac Lab improves on this with a partially object-oriented articulation API that allows one to create an articulation object (e.g., for a cabinet). However, one still has to often play around with index values to get the relevant articulation data they need. We use a cabinet opening task as a case study. In a cabinet drawer opening task, to write good reward functions you need to access the drawer link's handle mesh's pose, as well as the joint angle between the drawer and the cabinet. A visual comparison of the 3 APIs (simplified from the actual code) is shown in Figure \ref{fig:cabinet-code-comparison}.

Furthermore, pose information in ManiSkill3 is object-oriented and stored as batched Pose objects, enabling an easy to read, method chaining pattern of programming for working with poses. For the sake of an example, suppose that we have 2 poses $P_1$, $P_2$ and want to compute $(P_1 P_2)^{-1} P_1^{-1}$, ManiSkill3 provides a much simpler and method chainable API to do this compared to Isaac Lab, as shown in Figure \ref{fig:pose-code-comparison}.

\subsubsection{Robots and Controllers}
ManiSkill3 supports both URDF and Mujoco MJCF definition formats natively and builds articulated robots based on the URDF/MJCF directly. For each robot, ManiSkill3 further provides a number of pre-built configurable controller options for both GPU parallelized joint position control and inverse-kinematic (IK) control, modified from ManiSkill2 for GPU simulation. ManiSkill3 builds upon the PyTorch Kinematics package \citep{Zhong_PyTorch_Kinematics_2024} to support inverse-kinematic based controllers parallelized on the GPU, typically used to control robot arm end-effectors. These options are easily configured at runtime with either preset configurations or user-supplied controller configurations. Currently, there are 20+ different robots supported out of the box in ManiSkill3, a subset of which are visualized in Figure \ref{fig:robots-grid} in the Appendix. Finally, ManiSkill3 comes with \href{https://maniskill.readthedocs.io/en/latest/user_guide/tutorials/custom_robots.html}{extensive tutorials and examples} of how to tune and optimize robots for fast simulation, which has often proven a stumbling block for those new to robot simulation importing complex robots for the first time.

\subsection{Demonstration Datasets}

We leverage a variety of approaches to collect/generate our demonstration datasets infinitely at scale. For the simplest tasks, we write and open source some motion planning-based solutions to generate demonstration data. Some tasks with easy-to-define reward functions have dense reward functions defined and converged RL policies are used to generate demonstration data. For more difficult tasks, we collect demonstration data (typically about 10 demonstrations) via teleoperation tools. Then, we use RFCL \citep{DBLP:conf/iclr/TaoSC024-rfcl} or RLPD \citep{DBLP:conf/icml/BallSKL23-rlpd} to run fast online imitation learning and generate data from converged policies.

We further adapt the trajectory replay tool from ManiSkill2 to work with both CPU and GPU simulated demonstration data. The replay tool enables users to change the observations stored (e.g., state or rgbd, and allow modifying the rendering shaders used) as well as modifying the rewards stored (dense or sparse). In tasks involving the Franka robot arms, we provide code to convert actions from one controller type to another (e.g. joint position control to delta end effector pose control). Importantly we support collecting data in CPU/GPU simulation and replaying them in CPU/GPU simulation with different numbers of parallel environments via explicit control over randomization/RNG seeding at the per-parallel-environment level. This enables flexibility in trajectory replay as data collected on one machine with more GPU memory can be replayed on other machines with less GPU memory that cannot use as many parallel environments.

\section{Baselines and Results}

ManiSkill3 provides several popular robot learning baselines as well as simple reproducible setups for end-to-end trainable vision-based sim2real policies.

\subsection{Reinforcement Learning}

We provide two categories of RL baselines as follows:

\textbf{Wall-time Efficient Reinforcement Learning}: We include a torch based vectorized implementation of model-free RL algorithms PPO and SAC \citep{DBLP:conf/icml/HaarnojaZAL18-sac}, as well as the state-of-the-art model-based RL algorithm TD-MPC2 \citep{hansen2024tdmpc2}. Configurations for baselines are tuned to minimize training wall time with no regard to sample efficiency. Code for PPO and SAC is implemented based on CleanRL \citep{huang2022cleanrl} and leverages the torch compile and cudagraphs acceleration features introduced by LeanRL.

\textbf{Sample Efficient Reinforcement Learning}: All of the RL baselines in the wall-time efficient setting besides PPO are included here with configurations tuned towards more gradient updates and fewer environment steps to maximize sample efficiency.

We share results on training speed in Section \ref{sec:rl_results}. More in-depth details on RL setups and more training results are shared in Appendix \ref{appendix:rl_baselines}.

\subsection{Learning From Demonstrations (LfD) / Imitation Learning}
We provide two categories of LfD baselines as follows:

\textbf{Offline Imitation Learning}: We currently provide Behavior Cloning, Diffusion Policy \citep{DBLP:conf/rss/ChiFDXCBS23-diffusion-policy}, Action Chunking Transformer \citep{zhao2024aloha}, and PerACT \citep{shridhar2022peract} as baselines. We also support evaluating (but not training) several vision-language action (VLA) models, namely Octo \citep{octo_2023}, RT-X \citep{open_x_embodiment_rt_x_2023}, and RDT-1B \citep{liu2024rdt}. We leave to future work to support training VLA models on simulation data as RDT-1B has done with the previous ManiSkill2 datasets.

\textbf{Online Imitation Learning}: Online imitation learning generally refers to algorithms that learn from demonstrations in addition to collecting online environment transitions. We currently provide the two state-of-the-art baselines: Reinforcement Learning from Prior Data (RLPD) \citep{DBLP:conf/icml/BallSKL23-rlpd}, and Reverse Forward Curriculum Learning (RFCL) \citep{DBLP:conf/iclr/TaoSC024-rfcl}. Note that RFCL leverages simulation state resets.

\citet{DBLP:conf/corl/MandlekarXWNWK021-whatmattersinbc} show that imitation learning algorithm performance heavily depends on how the demonstrations were collected, particularly on how ``multimodal'' the data is. For example, some behavior cloning algorithms perform poorly when trained on motion planning or human teleoperated data, but can perform very well if trained on data generated by a deterministic neural network. With this caveat in mind, we explicitly track in all our LfD baselines the number of demonstrations used (an integer), what type of demonstrations are used (RL generated, motion planning, or human), and where the demonstrations are sourced from exactly (a longer description e.g. neural net trained via TD-MPC2, teleoperation via Meta-Quest VR). Imitation learning results on some environments are shared in Appendix \ref{appendix:il_baselines}.

\subsection{RL Training Speed}

\label{sec:rl_results}

\begin{figure}[h]
    \centering
    \includegraphics[width=\linewidth]{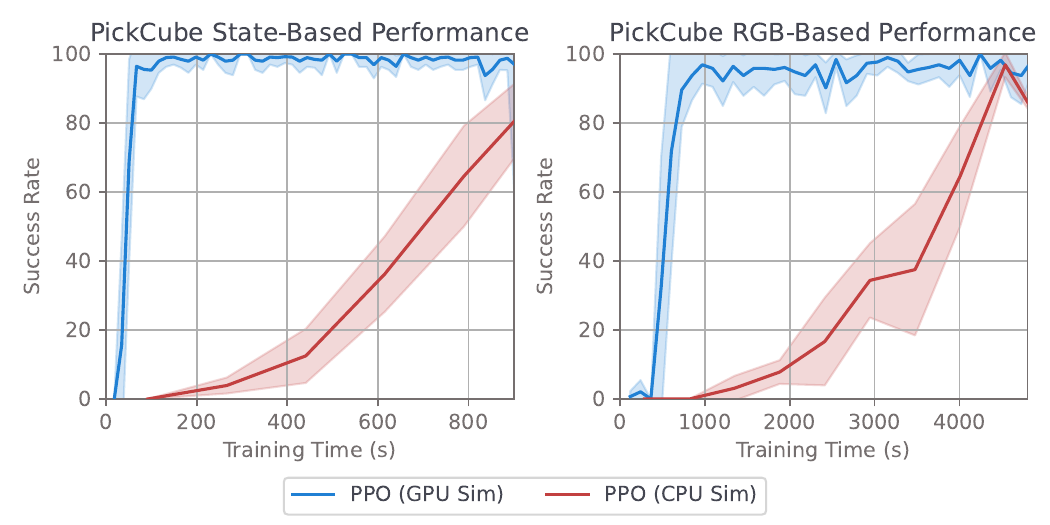}
    \caption{Wall-clock training time of PPO on GPU/CPU simulation showing the average success rate over time across 5 seeds. Shaded areas correspond to the 95\% confidence interval.}
    \label{fig:trainingspeed}
\end{figure}

We run experiments using PPO \citep{DBLP:journals/corr/SchulmanWDRK17-ppo} on the ManiSkill3 GPU simulation and the ManiSkill2 CPU simulation. ManiSkill2 was previously the fastest robotics simulation+rendering framework until ManiSkill3. The experiments were run on an RTX-4090 GPU on the PickCube task, where a Franka robot arm must grasp a randomly initialized cube and hold it still at a random goal location. For the vision-based task no ground truth data like cube pose is provided and the RL policy must solve from proprioceptive information and one 128x128 RGB image rendered by the environment's 3rd-person camera. RL hyperparameters are tuned to achieve the fastest training time in both settings. Results in Figure \ref{fig:trainingspeed} show that state and vision based training are massively accelerated with GPU simulation and rendering. PickCube with delta joint position control from state-based observations in GPU simulation reaches near 100\% success rate after about 1 minute of training, a 15x speed up relative to ManiSkill2. From RGB observations with parallel rendering PickCube is solved after about 10 minutes of training, a 8x speed up relative to ManiSkill2. For results and details of RL on more environments see Appendix \ref{appendix:rl_baselines}.

\subsection{Vision-Based Sim2Real Manipulation}

\begin{figure}[h]
    \centering
    \includegraphics[width=\linewidth]{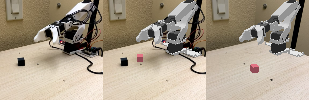}
    \caption{Green screening visualized with Koch-v1.1 robot. Left: Real RGB observations. Right: Sim RGB observations overlaid on real background. Middle: Sim RGB observations overlaid on real RGB observations for alignment visualization.}
    \label{fig:koch_greenscreen}
\end{figure}

\label{sec:sim2realvisionresults}

To showcase the use-case of heterogeneous simulation and parallel rendering, we provide a simple reproducible setup for end-to-end training a vision-based manipulation policy that deploys successfully zero-shot. The setup uses the low-cost \href{https://github.com/jess-moss/koch-v1-1}{\$300 Koch robot arm} and a phone camera for third-person RGB observations, leveraging the open-source Hugging Face LeRobot library \cite{cadene2024lerobot} for the hardware and setup to enhance reproducibility and accessibility. This setup is not limited to the Koch arm; other more expensive robotic arms can be used and likely perform better due to more precise hardware.

During simulation training and real-world evaluation, observations are restricted to RGB inputs and robot joint positions; no demonstrations or privileged state information such as cube pose is used, and the robot is controlled via continuous joint level input. We note however that future work combining simulation training with teleoperated demonstrations could improve performance and training time. To match the real background within simulation and reduce the visual sim2real gap we green screen the real-world background. We skip green-screening over the dynamic objects like the cube and robot arm via the environment's segmentation map, see Figure \ref{fig:koch_greenscreen} for an example. To enhance generalization and accommodate variations in real-world setups, we domain randomize a number of elements of the simulation. Task agnostic domain randomizations include camera pose, lighting direction, and the robot pose. Specific to the cube picking task we further domain randomize the cube size, color, and friction. 

In this demonstration we picked a random table in a house and set up the green-screening and roughly aligned the simulation camera pose with the real-world camera pose. Then we train on the cube picking environment with PPO using a simple Nature CNN backbone for image feature processing for 15 million samples using 256 parallel environments, taking approximately 1 hour on a RTX 4090 GPU. The trained policy is then zero-shot deployed on the real robot, using the same controller the policy was trained on in simulation, namely a delta target joint position controller. Evaluating the final checkpoint across 3 training runs in the real-world 8 times each yielded a $22/24 = 91.6\%$ success rate. Real world evaluations test on cubes of varying sizes, colors, and start poses, demonstrating our vision-based sim2real setup is capable of learning robust manipulation policies. Evaluation videos can be found on our \href{https://maniskill.readthedocs.io/en/latest/user_guide/demos/gallery.html#vision-based-zero-shot-sim2real-manipulation}{demo gallery page}.

\begin{figure}[h]
    \centering
    \includegraphics[width=\linewidth]{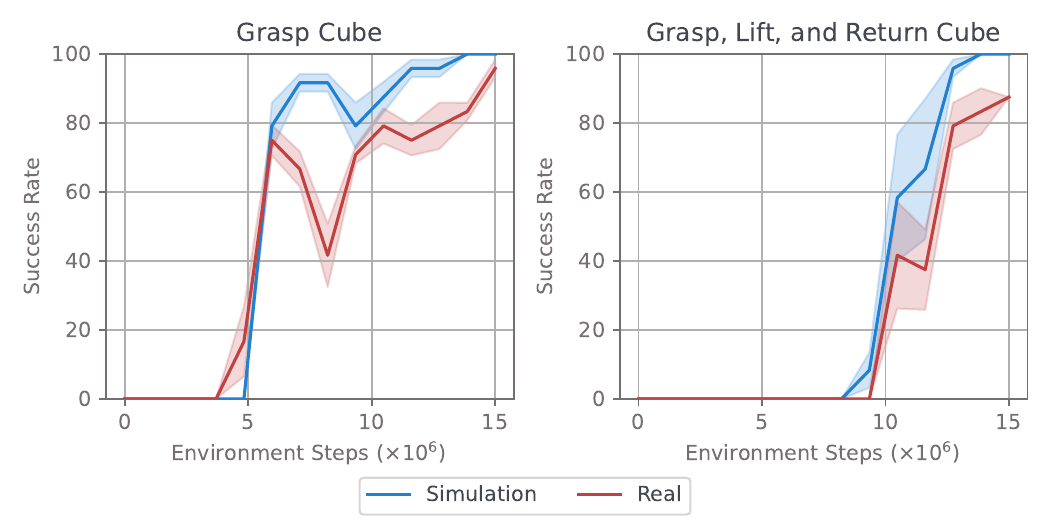}
    \caption{Koch pick-cube sim and real success rates on the grasp cube subtask as well as the full success consisting of grasping, lifting, and return the cube to a goal position. 14 training checkpoints for each of the 3 seeds are evaluated on both sim and real 8 times, using a variety of cube colors and sizes. Success rates are averaged across trials and shaded areas represent 95\% confidence intervals.}
    \label{fig:koch_sim2real_results}
\end{figure}

We further evaluated each intermediate checkpoint from the 3 RL training runs and plot the simulation and real-world success rates in Figure \ref{fig:koch_sim2real_results}. The figure shows a good correlation between real-world and simulated success rates, indicating that some sim2real digital twins built with ManiSkill can be fairly accurate and reflect the real-world. Interestingly, we observe that despite not modelling shadows or the exposed wiring of the robot arm accurately, a successful sim2real policy is still capable of being trained. See Appendix \ref{appendix:sim2realsetup} for more details on the exact setup, domain randomizations, controller implementation, and more.

\section{Limitations}

While parallel rendering can enable some new and more accessible approaches to sim2real and visual RL, it has its limits. For complex environments with many geometries (e.g. room-scale scenes) we can GPU parallel render and simulate them but at a lower number of environments on one GPU.

While it is now possible to do some more generalizable zero-shot RGB based sim2real with a relatively simple setup, some reward engineering was required to encourage safer and more robust grasp behaviors of RL trained policy. Moreover, this is far from solving sim2real completely, but more of opening a new avenue of potential approaches based on fast visual RL in simulation. The sim2real demo is also limited to using just static cameras and it can be interesting future work to explore how to use methods like Gaussian Splatting for novel-view rendering to support mounted cameras. One could also explore algorithnmic changes and/or domain randomizations that can help remove the need for green screening.

Not all types of environments in ManiSkill are readily "GPU parallelized" in the sense that there are a batch of parallel environments. Most GPU parallelized environments are rigid-body based. The soft body environments are not batched as they use a significant portion of the GPU to simulate just a single environment fast. The vision-tactile simulator that has been tuned heavily for sim2real transfer uses a different set of algorithms for physics simulation compared to the majority of rigid-body only environments.

\section{Conclusion}
\label{sec:conclusion}
ManiSkill3 introduces a state-of-the-art framework/benchmark for generalizable robotics simulation and rendering. ManiSkill3 uses less GPU memory and depending on scenario can run faster while also supporting heterogeneous GPU simulation. Additionally we support the most diverse range of robotics tasks compared to alternative simulators. Importantly, we reliably support both sim2real and real2sim environments in manipulation tasks with real-world reproducible results. Moreover, we provide a immersive VR teleoperation system option for users to collect data for their own research. Furthermore, ManiSkill3 provides an easy-to-use object-oriented API for building all kinds of GPU simulated tasks, democratizing access to scalable robot learning. Finally, demonstrations and RL/IL baselines with clearly defined metrics are open sourced for users to use. We believe that our comprehensive approach to building the open-osurce ManiSkill3 will encourage the research community to tackle manipulation challenges more extensively through compute-scalable simulation.

\section*{Acknowledgments}
The authors sincerely thank members of the Hao Su lab and Hillbot for great discussion and feedback, including but not limited to Nicklas Hansen, Rokas Bendikas, Jiayuan Gu, James Hou. The authors also thank members of the open source community for bug finding, contributions, and feedback on this framework, including but not limited to Alexandre Brown, Gaotian Wang, Dwait Bhatt, Jeremy Morgan. We especially acknowledge the generous support from Qualcomm Embodied AI Fund and Hillbot Inc. Stone Tao is supported in part by the NSF Graduate Research Fellowship Program grant under grant No. DGE2038238.

\bibliographystyle{plainnat}
\bibliography{references}
\clearpage
\textbf{Appendix Table of Contents}
\begin{itemize}
    \item Environments and Robots \ref{appendix:tasks}
    \item Reinforcement Learning Baselines \ref{appendix:rl_baselines}
    \item Imitation Learning Baselines \ref{appendix:il_baselines}
    \item Simulation and Rendering Benchmarking \ref{appendix:sim-render-benchmark}
    \item VR Teleoperation \ref{appendix:teleop}
\end{itemize}
\section{Environments and Robots List}
\label{appendix:tasks}

\begin{figure}[ht]
    \centering
    \includegraphics[width=\linewidth]{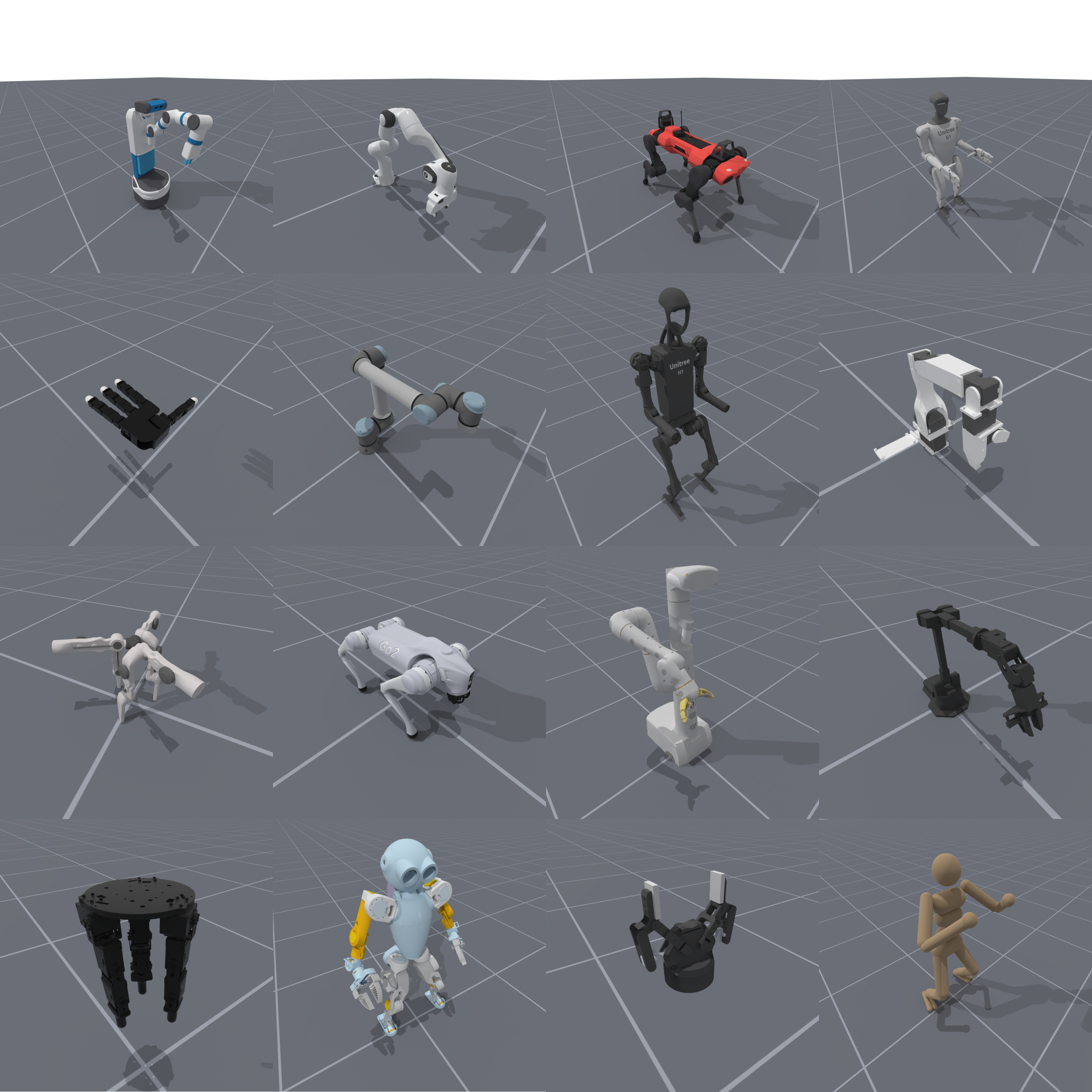}
    \caption{A sample of 16 of the robots supported in ManiSkill3 on both CPU and GPU simulation.}
    \label{fig:robots-grid}
\end{figure}

This section covers key implementation and optimization details of the general supported task categories. For a complete list with videos of almost every task, see \href{https://maniskill.readthedocs.io/en/latest/tasks/index.html}{ManiSkill3 Tasks Documentation}. For videos of soft body and vision-tactile tasks we refer readers to the ManiSkill2 paper \citep{} and A sample of 16 of the supported robots, which include mobile manipulators, floating grippers, quadrupeds, etc. is displayed in Figure \ref{fig:robots-grid}. For a complete list with details/pictures of nearly every robot, see \href{https://maniskill.readthedocs.io/en/latest/robots/index.html}{ManiSkill3 Robots Documentation}.

We emphasize here that while ManiSkill3 may not have the most distinct number of tasks compared to some benchmarks, the core contribution is supporting a diverse array of possible tasks with open-sourced code that users can reference and use to easily build more tasks.

\subsection{Table Top Manipulation}
\begin{figure}[h]
    \centering
    \includegraphics[width=\linewidth]{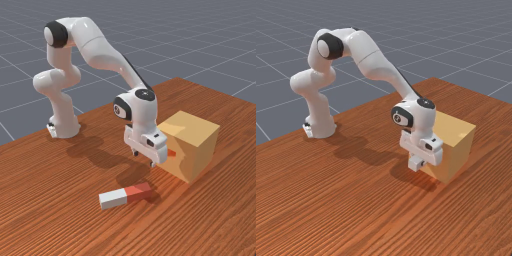}
    \caption{Example table-top manipulation task showing the start state and the solved state. The robot arm must grasp the white-orange peg and insert it into the box.}
    \label{fig:peginsertionside}
\end{figure}
Table-top manipulation is primarily related to controlling one or more robot arms to manipulate an object on a table. Robots like Franka Emika Panda and Universal Robots 5 fall under this category. Typical tasks may include picking up objects, inserting a peg, assembling a structure, pushing objects, etc. An example task is shown in Figure \ref{fig:peginsertionside}.

\textbf{Implementation Details:} All robot arms are modified to have certain impossible self-collisions disabled and some have their collision meshes modified for faster simulation. 

\subsection{Mobile Manipulation}
\begin{figure}[h]
    \centering
    \includegraphics[width=\linewidth]{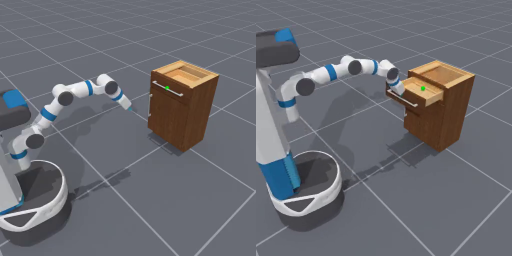}
    \caption{Example mobile manipulation task showing the start state and the solved state. The mobile robot arm must grasp the handle indicated by the green sphere and pull it open.}
    \label{fig:opencabinetdrawer}
\end{figure}
Mobile manipulation here refers to tasks in which a robot arm has a mobile base. Robots like Fetch and Stretch fall into this category. Typical tasks may include placing objects on surfaces, opening cabinet doors/drawers, picking up objects off the ground, etc. Example task is shown in Figure \ref{fig:opencabinetdrawer}.

\textbf{Implementation Details:} The default robot supported is Fetch. The mobile base in particular is not simulated by driving the wheels, and is modeled similarly to AI2-THOR \citep{ai2thor} and Habitat \citep{szot2021habitat} with one joint controlling forward/backward movement and another controlling rotation of the base. The Fetch robot definition and collision meshes have further been tuned to be simpler for faster simulation. Several impossible self-collisions between some links have been explicitly ignored to speed up simulation.

\subsection{Room Scale Environments}

ManiSkill3 provides out-of-the-box code to build the ReplicaCAD environment from Habitat \citep{szot2021habitat}, all AI2-THOR environments \citep{ai2thor} using assets compiled by the authors of the Habitat Synthetic Scenes Dataset \citep{khanna2023hssd}, and the RoboCasa scenes dataset \citep{robocasa2024}. Photo-realism is also possible by turning on the ray-tracing shader options when creating the environment.

AI2-THOR \citep{ai2thor} and Habitat \citep{szot2021habitat} have long-horizon mobile manipulation tasks but rely on slow CPU simulation / slow rendering systems, and do not support contact-rich physics for manipulation (only magical grasp) and do not look photorealistic. Robocasa \citep{robocasa2024} has contact-rich long-horizon mobile manipulation tasks in photorealistic room-scale environments. However, Robocasa simulates and renders these scenes at around 25FPS as it does not use GPU parallelized simulation and rendering. In contrast ManiSkill3 can simulate the complex ReplicaCAD environment up towards 2000+ FPS with rendering. 

\textbf{Implementation Details:} We further make several modifications to ReplicaCAD to make it completely interactive as some of the collision meshes for articulations were modelled incorrectly and thus did not support low-level grasping. Via CoACD \citep{wei2022coacd} we run convex decomposition on objects in AI2-THOR scenes to generate simulatable non-convex collision meshes so those objects can e.g. be grasped and moved around correctly. Via manual annotation by ManiSkill3 authors, certain categories of objects in AI2-THOR are made to be kinematic so they cannot be moved around (e.g. tables, TVs, clocks, paintings) with the rest allowed to be dynamic to be fully simulated (e.g. apples, baseball bat, cups) to optimize simulation speed.

\subsection{Locomotion}
\begin{figure}[h]
    \centering
    \includegraphics[width=\linewidth]{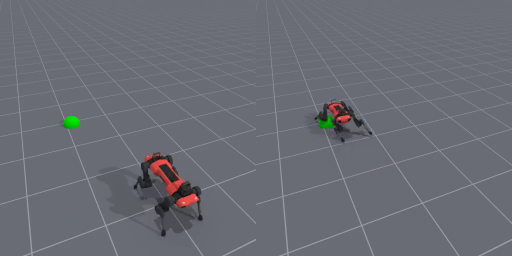}
    \caption{Example locomotion task showing the start state and the solved state. The quadruped must use its 4 legs to to move to the green goal.}
    \label{fig:anymalcreach}
\end{figure}
Locomotion here refers to controlling robot joints to move a robot from one location to another. Quadrupeds such as AnyMAL-C and humanoids such as Unitree-H1 fall into this category. Example task is shown in Figure \ref{fig:anymalcreach}.

\begin{figure}[h]
    \centering
    \includegraphics[width=\linewidth]{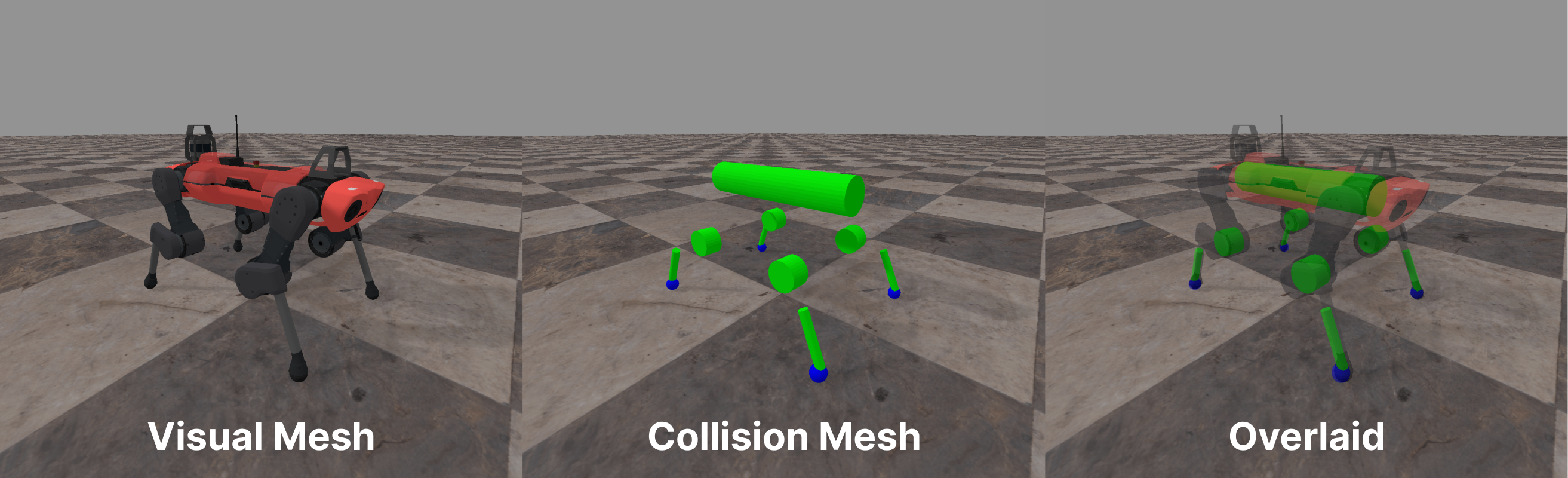}
    \caption{Comparison of the visual and collision mesh of one of the robot quadruped models, AnyMAL-C.}
    \label{fig:anymal-c-collision-mesh}
\end{figure}

\textbf{Implementation Details:} Similar to Isaac Lab, quadruped robots in locomotion tasks are modified such that the large majority of collision meshes are removed, leaving behind just the feet, ankles, and the body visualized in Figure \ref{fig:anymal-c-collision-mesh}. Moreover following Isaac Lab, joint limits are significantly constrained such that random actions do not easily cause the robot to fall over. 

\subsection{Humanoid / Bi-manual Manipulation}
\begin{figure}[h]
    \centering
    \includegraphics[width=\linewidth]{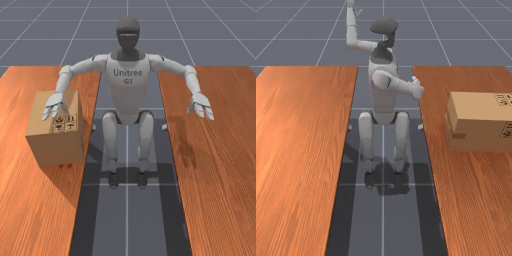}
    \caption{Example humanoid manipulation task showing the start state and the solved state. The humanoid must use both arms to grasp the box and transport it to the other table.}
    \label{fig:unitreeg1-transportbox}
\end{figure}
Tasks here refer to the use of a humanoid embodiment such as the Unitree H1 robot or bi-manual robot embodiments for manipulation tasks. Example task shown in \ref{fig:unitreeg1-transportbox}.

This type of task can be found in benchmarks like RoboCasa and BiGym, although we note that RoboCasa and BiGym do not reliably support working RL setups and are not GPU parallelized. Isaac Lab currently does not have these types of tasks out of the box.

\textbf{Implementation Details:} For some tasks where the robot has legs, to simplify the task the legs are fixed in place so that robot learning methods can focus on manipulation only and train faster. Tasks still have the option to swap a version of the robot where all joints are controlled although they are much harder.

\subsection{Multi-Agent Robots}
\begin{figure}[h]
    \centering
    \includegraphics[width=\linewidth]{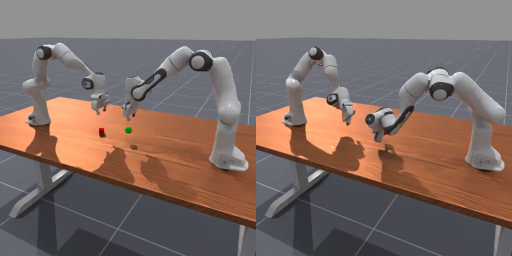}
    \caption{Example multi-agent robotics task showing the start state and solved state. The left robot must push a cube over to the other robot to pick up.}
    \label{fig:tworobotpickcube}
\end{figure}
Multi-Agent robots refer to support for controlling multiple different robots in the same simulation to perform a task. Setups such as multiple quadrupeds or robot arms fall into this category. A common task is the handover of objects. Example task shown in Figure \ref{fig:tworobotpickcube}.

This type of task can be found in Robosuite. Isaac Lab supports this type of task, but not out of the box. Past environments based on older versions of Isaac have example tasks with multiple robot arms/hands running on GPU simulation \citep{chen2022towards}.

\textbf{Implementation Details:} By default the action space is a dictionary action space in multi-agent environments with a dictionary key for each controllable agent, which follows the standard PettingZoo API \cite{terry2021pettingzoo}. PettingZoo is currently the most popular interface for multi-agent RL environments. For users who do not wish to do multi-agent RL they can flatten the action space into a single vector if necessary via a environment wrapper provided by ManiSkill3.

\subsection{Drawing/Cleaning}
\begin{figure}[h]
    \centering
    \includegraphics[width=\linewidth]{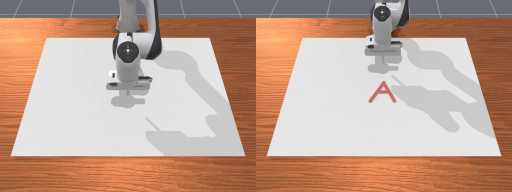}
    \caption{Example drawing/cleaning task showing the start state and solved state. The robot must draw a given SVG file (which describes a set of lines), in this case the letter A.}
    \label{fig:drawlettersvg}
\end{figure}
Drawing/Cleaning refers to tasks for dynamically "adding/removing" objects to simulate the effect of drawing or cleaning. A task could be to draw the outline of a shape on a canvas or clean dirty spots on a table surface. Example task shown in Figure \ref{fig:drawlettersvg}.

ManiSkill3 is the only framework that supports this kind of task out of hte box with GPU parallelization and rendering.

\textbf{Implementation Details:} The drawing/cleaning effect is achieved by building ahead of time 1000s of small thin cylinders that represent ``ink" or dirty spots. For a drawing task, all of these cylinders are hidden away from the camera view. When a robot moves a drawing tool close to a surface/canvas, the cylinders have their pose set to be on top of the surface right under where the drawing tool is. For a cleaning task, all the cylinders/dirty spots are visible and removed once the cleaning tool moves over the dirty spot. Currently the drawing/cleaning environments in ManiSkill3 do not require intricate grasping (nor is it the focus) of the drawing/cleaning tool, so the solver position/velocity iteration values are tuned down.

\subsection{Dextrous Manipulation}
\begin{figure}[h]
    \centering
    \includegraphics[width=\linewidth]{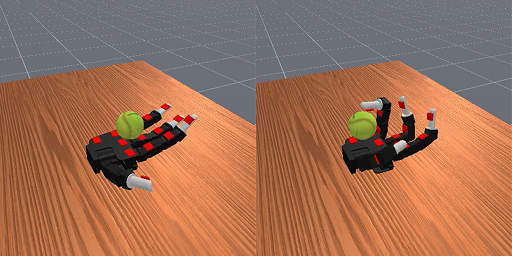}
    \caption{Example dextrous manipulation robotics task showing the start state and solved state. The robot hand must rotate the object in place to a desired orientation.}
    \label{fig:rotatesingleobject}
\end{figure}
Dextrous Manipulation refers to tasks often involving multi-fingered hands and dense/rich contacts occurring during manipulation. An example task is in-hand rotation. Example task shown in Figure \ref{fig:rotatesingleobject}.

This type of task has been heavily explored and exists in Isaac Lab and ManiSkill3 with GPU parallelization.

\textbf{Implementation Details:} Not many special optimizations are made here. Tactile sensing is further provided in environments via touch sensors placed on the dextrous hand at various points.

\subsection{Vision Tactile Manipulation}
\begin{figure}[h]
    \centering
    \includegraphics[width=\linewidth]{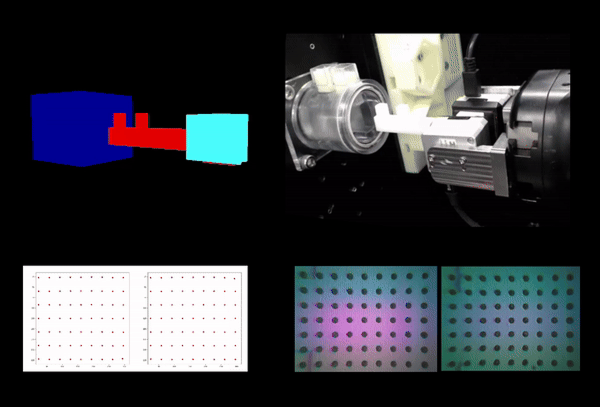}
    \caption{Example task of a key insertion task in simulation (top left), the real world equivalent (top right), and plots of the tactile feedback data (bottom).}
    \label{fig:vitac-keyinsert}
\end{figure}
Vision Tactile Manipulation refers to tasks that rely on processing tactile information like images to solve manipulation tasks. Tasks could include key insertion which require tactile inputs to solve due to visual occlusions. Example task shown in Figure \ref{fig:vitac-keyinsert}

\textbf{Implementation Details:} The vision-tactile, sim2real, manipulation environments are ported over from the 2024 ManiSkill Vision-based-Tactile Manipulation Skill Learning Challenge \citep{chen2024tactilesim2real}.

\subsection{Classic Control}
\begin{figure}[h]
    \centering
    \includegraphics[width=\linewidth]{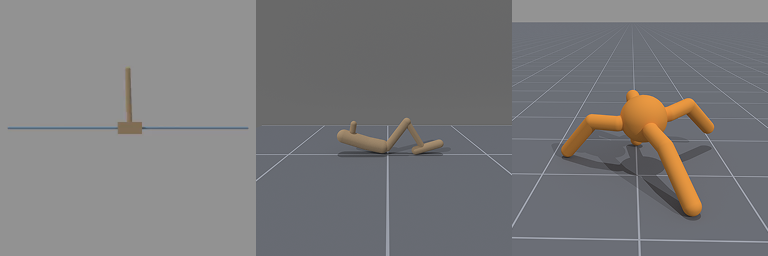}
    \caption{Example control environments: CartPoleBalance, Hopper, and Ant.}
    \label{fig:mscontrolenvs}
\end{figure}
Classic control is quite broad but in this context refers to fake robots tasked with achieving some stable pose or moving in a direction at desired speeds. Tasks include cartpole balancing, hopper etc. Examples are shown in Figure \ref{fig:mscontrolenvs}

DM-Control \citep{DBLP:journals/simpa/Tunyasuvunakool20-dmcontrol} has the most implemented control tasks and Isaac Lab has a few GPU parallelized variants. ManiSkill3 has GPU parallelized simulation+rendering variants of the most control tasks.

\textbf{Implementation Details:} ManiSkill3 uses its MJCF loader to load the MJCF robot definitions from the original DM-Control repository and tunes robot pd joint delta position controllers to align as closely as possible to the behavior seen in DM-Control/Mujoco.

\subsection{Digital Twins}
\label{appendix:digitaltwins}
Digital twins have two variants included, environments for real2sim and environments for sim2real. The distinction here is real2sim environments simply need to be designed so that a model trained on a real world equivalent of the simulation environment achieves similar success rates when evaluated in simulation. Sim2real digital twins are environments designed so that models trained on simulation data can be more easily used for real world deployment.

\begin{figure}[h]
    \centering
    \includegraphics[width=\linewidth]{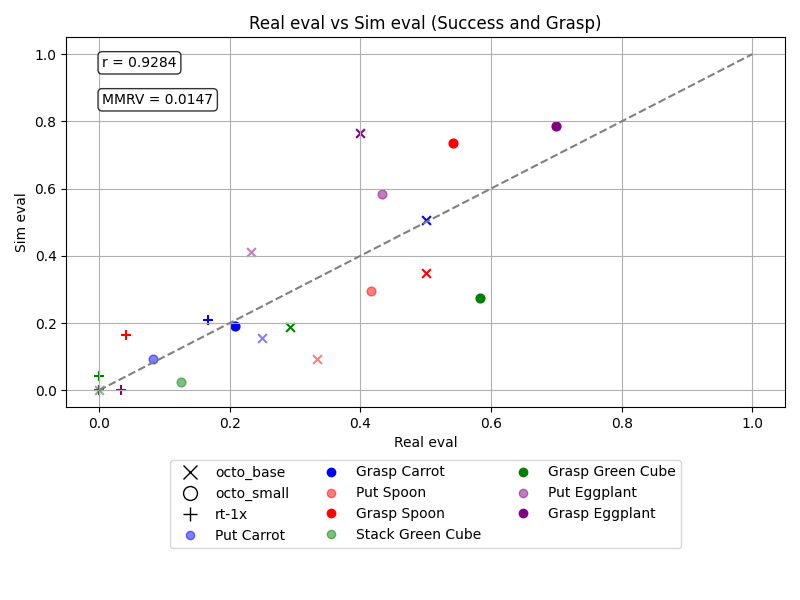}
    \caption{Evaluated success rates of generalist robotics models like Octo and RT-1X on 4 different tasks. The correlation and MMRV metrics are close to that of the original paper. MMRV is Mean Maximum Rank Violation (lower the better), which assess the accuracy of the consistency of the rankings of real and sim policy evaluations.}
    \label{fig:simpler_real_sim_eval}
\end{figure}

For real2sim, ManiSkill3 ports over and GPU parallelizes some environments from SIMPLER \citep{li24simpler}, which enables efficient evaluation of policies trained on real world data like the generalist RT-X and Octo models. The primary tricks include green-screening a real world image and texture matching which have been copied over and parallelized. We ensured the GPU parallelized port of SIMPLER achieves similar results/behaviors as the original CPU simulated environments as shown in Figure \ref{fig:simpler_real_sim_eval}.

For sim2real, we provide tools useful for training policies in simulation and directly deploying into the real-world. Given the number of details involved in sim2real, we explain the implementation details of vision-based sim2real in detail in Appendix \ref{appendix:sim2realsetup}.

\subsection{Soft Body Manipulation Environments}
Soft body manipulation refers to the manipulation of soft body objects that can deform and morph in shape. Tasks include excavating sand particles, pouring water etc. 

ManiSkill3 soft body manipulation environments are the same as ManiSkill2 which uses 2-way coupled rigid-MPM simulation that enables rigid body objects to interact with soft body objects. We point readers to the ManiSkill2 paper for more details on soft body simulation.

\clearpage

\section{Vision-Based Sim2Real}
\label{appendix:sim2realsetup}

We describe how one can use ManiSkill3 to perform end-to-end vision-based sim2real by training with PPO a RGB-based manipulation policy in simulation and zero-shot deploying it in the real world. Past work such as TRANSIC-Envs \citep{jiang2024transic} provide reproducible sim2real digital twin setups but rely on state estimation via ArUco markers for sim2real transfer and does not support visual feedback. Dextreme \citep{DBLP:conf/icra/HandaAMPSLMWZSN23-dextreme} provides a realistic in hand cube rotation environment but also does not support efficient visual feedback and relies on accurate state estimation. ManiSkill3 on the other hand has fast visual data generation that enable mimicing real-world cameras inputs for training at scale and fast. We note however the example setup here is a task less complex than that of e.g. Dextreme but point out that our reproducible setup is easily extendable and could be the start of opening a new avenue of sim2real research with efficient and fast large-scale simulation training on image inputs.

\subsection{Hardware}

We use the low-cost \$300 Koch v1.1. robot arm and the LeRobot library for a simple accessible python interface to control the robot arm. Due to the simple but accessible hardware, it is generally harder to deploy policies on this robot as the motors are less precise compared to more expensive robot arms. Thus the vision-based sim2real demonstrated here can easily work on other robot arms. Apart from the robot we also setup one third view iPhone camera (any LeRobot compatible camera can work) to record RGB observation data.

\subsection{Controller / Action Space}

We use a target delta joint position controller in both simulation and the real-world at a control frequency of 30Hz. At timestep $t=0$ after an (real or simulation) environment reset we save an initial target joint position value $\bar{q}_t = q_t$ where $q_t$ is the joint position of the robot arm at timestep $t$. Given an action $a_t$ at timestep $t$, we set $\bar{q}_{t+1} = \bar{q}_t + a_t$.

At each timestep $t$, we drive the robot's joint positions to $\bar{q}_t$. In simulation this is equivalent to setting target joint position values in the PhysX physics engine and the behavior is defined by a PD controller. In the real world this is equivalent to setting the next joint position of the robot arm to be $\bar{q}_t$, and most robot arms have APIs that enable this functionality and will internally run their own control loop to reach the set joint position.

The actual action space of the environment is normalized to a range of -1 to 1 for each joint. Given a predicted action from a policy $\hat{a}_t$ we clip it first to the range of $[-1, 1]$ and then unnormalize it to obtain $a_t = \frac{\hat{a}_t (h-l)}{2}$, which is the delta action applied to the $\bar{q}_t$. For the Koch robot arm $l=-0.05, h=0.05$, which represent a maximum change of 0.05 radians in either direction of each joint.

We choose the target delta joint position controller as the sim2real dynamics gap for quasi-static tasks (tasks where objects only move when the robot manipulate them) is minimized. In both simulation and real-world in quasi-static settings with small enough action sizes as long as within the timeframe of one control step the real world robot and simulated robot reach $\bar{q}_t$ at timestep $t$ then the simulation and real-world are aligned well. We note that we leave it to future work to explore controller and sim2real design for addressing more dynamic tasks such as catching a thrown object which require higher control frequencies.

\subsection{Observation Space}
The observation space follows that of all visual RL baselines, which contains both proprioceptive data and the camera input RGB data. For the simulated tasks the prioprioceptive data contains both the current robot joint positions $q_t$ as well as the target joint positions $\bar{q}_t$ which is required to make the problem markov and more easily solvable via RL. The RGB data is a $128 \times 128$ RGB image captured during each control step of the environment. 

We further include a binary label of whether or not the robot is grasping something, which is a simple check of whether the joint position of the gripper joint is within 0.02 radians of the target joint position of the gripper joint. If the joint position of a robot cannot reach the target joint position this indicates something is blocking that joint, e.g. an object being grasped. Empirically we find that due to occlusions that occur once the cube is grasped due to the camera angle and the gripper link covering the cube, the trained policy has difficulty discerning whether or not it has grasped a cube. The binary is grasped feature generated from $q_t, \bar{q}_t$ values is sufficient to overcome the issue of occlusions.

All of these observation data points are easily obtained in the real world without needing to set up perception stacks to observe e.g. object poses. No observation delays are added but can be useful for higher frequency tasks.

\subsection{Environment Setup and Domain Randomizations}
\begin{figure}[!th]
    \centering
    \includegraphics[width=\linewidth]{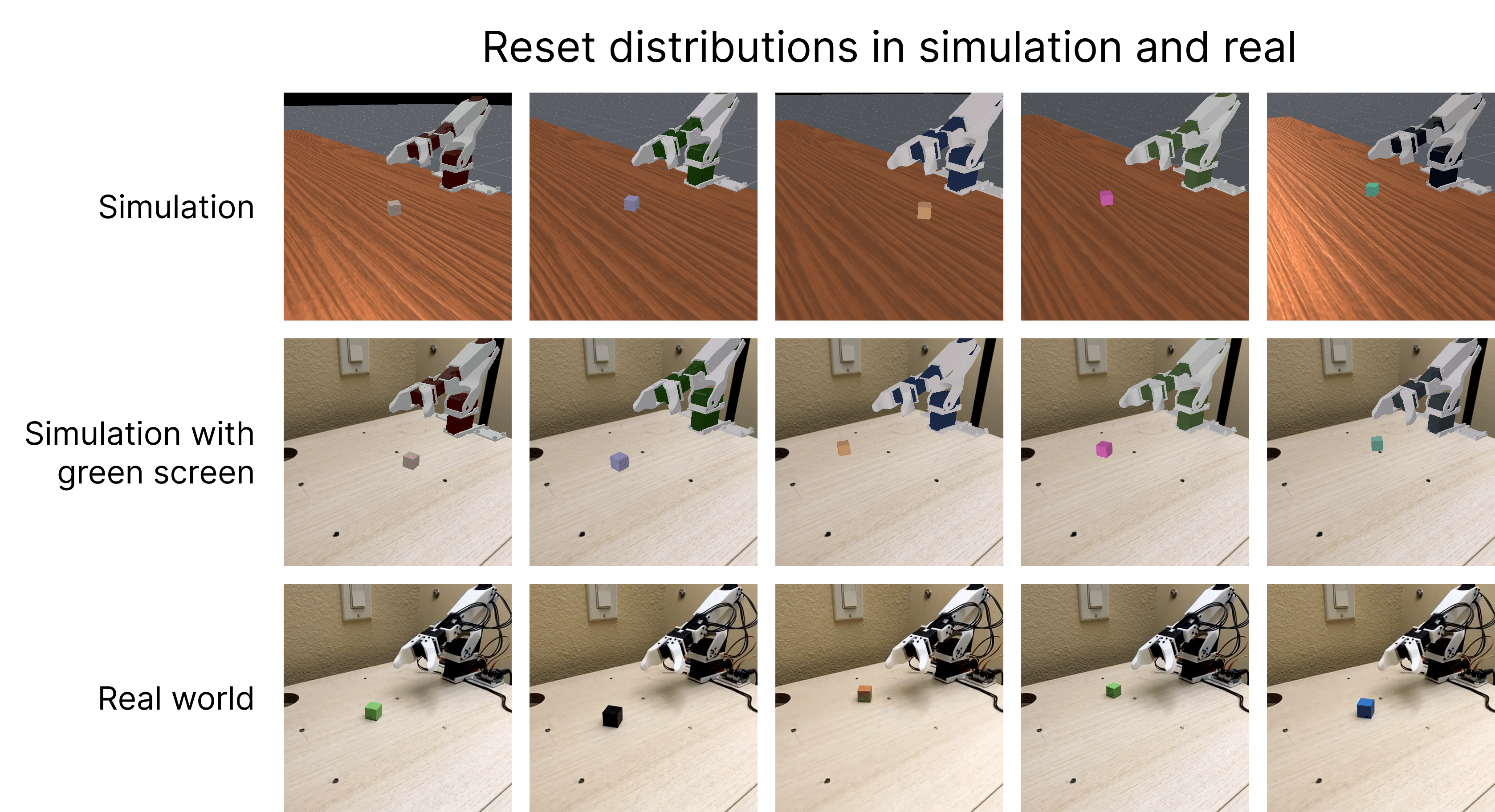}
    \caption{Sample images of the start states of the simulated environment and real-world environments, showing the reset distribution. Videos of the reset distribution can be found on our website.}
    \label{fig:koch_sim2real_reset_distribution}
\end{figure}
The simulated environment contains 4 objects, the floor, the table, the cube, and the robot arm. A visualization of the reset distribution can be found in Figure \ref{fig:koch_sim2real_reset_distribution}.
We perform the following randomizations at each timestep of the environment:
\begin{itemize}
    \item Camera pose is randomized by randomly shifting the camera position up to 2.5 cm in x, y, and z axes in the world frame. This randomization scale sufficiently covers possible misalignments between simulated and real-world cameras.
\end{itemize}

We perform the following randomizations each time the environment is reset:
\begin{itemize}
    \item The initial robot joint positions $q_0$ are sampled from $\mathcal{N}(q_R, 0.02)$, where $q_R$ represents the joint positions we set for the real Koch v1.1 robot at each reset.
    \item The position of the cube is uniformly sampled within a $10$cm$^2$ square in front of the robot arm, and the cube's z-axis rotation is randomized completely.
\end{itemize}
We perform the following randomizations each time the environment is reconfigured, which permit the environment to use different object assets, materials, camera settings. This occurs only during environment resets, but typically is done less often as it slows down reset times.
\begin{itemize}
    \item Cube side lengths are sampled uniformally in the range $[1.5, 2.25]$ cm
    \item Cube frictions are sampled from $\mathcal{N}(0.3, 0.025)$
    \item Cube color is sampled uniformally from all possible RGB values.
    \item Robot chassis color is sampled from $\mathcal{N}(c, 0.05)$, where $c$ is the base robot chassis RGB color scaled from 0 to 1.
    \item Robot motor color is sampled from $\mathcal{N}(m, 0.05)$, where $m$ is the base robot motor color scaled from 0 to 1.
\end{itemize}

\subsection{Reward}
$\mathbb{I}_{grasped}$ indicates whether the robot is grasping the cube. $\mathbb{I}_{touch\_table}$  indicates if the robot is robot's gripper is colliding with the table. $\mathbb{I}_{tcp\_close}$ indicates whether the tool center point, the point between the robot's two gripper fingers $x^{tcp}_t$, is within a half of the cube's side length from the cube's position, $x^{cube}_{t}$. We also define the vector from the first gripper finger to the second as $\vec{g}$ and the cube size as $c$. The reward function, for a state $s$ and action $a$ is: 
\begin{align*}
r_t(s,a) = &\ r_{dist}(x^{tcp}_t, x^{cube}_{t}, 15) + r_{orient}(\vec{g}) \\  
           &+ 2\mathbb{I}_{tcp\_close} r_{dist}(||\vec{g}||, c, 40) \\  
           &+ \mathbb{I}_{grasped} \left( 1 + 3r_{dist}(q_t, q_R, \frac{4}{\pi}) \right) \\  
           &- 2\mathbb{I}_{touch\_table}  
\end{align*}

$r_{dist}$ is a common distance reward function used in ManiSkill3, where $r_{dist}(x,y,z) = 1 - tanh(z * ||x-y||)$. $r_{orient}(\vec{g}) = 1 - (\vec{u} \cdot \frac{\vec{g}}{||\vec{g}||})$, where $\vec{u}$ is the up vector. Intuitively, this means rewarding the agent with a  value of one for keeping $\vec{g}$ perpendicular to the table normal and a value of zero if it is parallel instead. $q_t$ and $q_R$ represent the joint positions at timestep $t$ and upon environment reset, respectively.

We find that action magnitude, $||a||$, does not need to be considered within the reward, as the controller's restrictions of  $l$ and $h$ are appropriately tuned for safe deployment and for the realistic maximum joint acceleration magnitudes for each real robots' motors. 

\subsection{Training}

PPO with RGB inputs is used for training, details on the overall RL implementation is discussed in Appendix \ref{appendix:rl_baselines}. Specific to this task we use a reconfiguration frequency of 10 to ensure the RL policy sees a sufficiently diverse set of randomizations over object geometry and materials while also ensuring we are not performing slow reconfiguration resets too often. Ultimately training takes about 1 hour on a 4090 GPU. The final training curves can be found in the main paper in Figure \ref{fig:koch_sim2real_results}.

\clearpage
\section{Reinforcement Learning Baselines}
\label{appendix:rl_baselines}

We most comprehensively test PPO and TD-MPC2 as reinforcement learning baselines, which represent two ongoing sides of RL research, walltime efficient RL and sample efficient RL. SAC is also provided but is not heavily benchmarked as of now. When possible RL baselines have sample-efficient and wall-time efficient hyperparameters. These baselines are implemented with PyTorch \citep{Ansel_PyTorch_2_Faster_2024}.

\subsection{Proximal Policy Optimization (PPO)}

\textbf{Neural Network Architecture}:
For state based tasks the actor and critic networks are separated and are comprised of a MLP with three 256 unit hidden layers followed by Tanh activation. For vision based tasks the actor and critic share a feature processing backbone, namely a NatureCNN following the default setup that the popular Stable Baselines 3 RL library \citep{stable-baselines3} uses. 

\textbf{Training Details and Hyperparameters}: 
The key hyperparameters that were tuned per environment were largely the discount factor, GAE lambda factor, the max episode steps/horizon, as well as the number of parallel environments. Typically we try to use up to 4096 parallel environments when possible for state based training and about 256 to 1024 environments for RGB based training. Discount factor by default is kept at a low value of 0.8 which empirically works well for most tasks and increased up to 0.99 if 0.8 does not work. Episode horizon is tuned such that the task can be solved within that horizon and the horizon is not much longer than the optimal solve time. We further tune the number of steps sampled from each parallel environment in each rollout. Generally a value of 16 or 32 is used and is tuned down if possible. However, we observe most tasks cannot be solved with low number of steps sampled per parallel environment like PegInsertionSide, likely due to long horizon temporal dependencies that are harder to model.

For all experiments we always turn partial resets on, which enables environments to reset early upon reaching a fail or success state if there are any. Partial resets generally enable faster training by reducing the time spent sampling the environment in irrelevant states.

ManiSkill environments support setting environment reconfiguration freqencies, which control how often during an environment reset we will also permit resetting/randomizing simulation properties that are not changeable after starting simulation (such as the object shape as done in the PegInsertionSide task). By default all experiments have a reconfiguration frequency of 0 for training environments as generally with sufficiently high number of parallel environments there is enough diversity in the training data. For more advanced use-cases like training robust sim2real policies as done in Appendix \ref{appendix:sim2realsetup}, we use a non-zero reconfiguration frequency to better randomize the object geometries and other properties.

Due to the number of hyperparameters and environments we do not include them in the paper itself and refer users to the public training runs on the \href{https://wandb.ai/stonet2000/ManiSkill/reports/PPO-Results--VmlldzoxMDQzNDMzOA}{Wandb page for PPO (state and RGB) results} which show all metrics and configurations.

\subsection{Temporal Difference Learning for Model Predictive Control 2 (TD-MPC2)}
\textbf{Neural Network Architecture}: For tasks with state-based observations, we used the original TD-MPC2 architecture from the original paper. For vision-based tasks, we modified the encoder to handle larger 128×128 input images and additional external state information (e.g., joint position, goal position). In particular, we introduced an additional convolutional layer to the RGB encoder, as well as a separate MLP (composed of one 256-unit and one 512-unit layer-normed, linear layer) to process the extra state data. The outputs from the RGB encoder and state encoder are then combined via a 512-unit linear layer. All other network components (e.g., actor and critic) remain unchanged from the original implementation.

\textbf{Training Details and Hyperparameters}:
We evaluate TD-MPC2 with the wall-time-efficient mode (using 32 parallel GPU environments). Due to the large number of hyperparameters and environments, we do not list them all here. Instead, we refer readers to our public training runs on the \href{https://wandb.ai/stonet2000/ManiSkill/reports/Off-Policy-RL-Results--VmlldzoxMTE2ODk2NA}{Wandb page for TD-MPC2 (state and RGB) results}, which include full metrics and configurations.

We only tune two parameters, the replay buffer size and the environment control mode. For all vision-based tasks, we use a smaller replay buffer size of 200K, while for state-based tasks we use 1M. Generally, we employ the delta joint position control mode for wall-time-efficient runs as the environment runs faster relative to IK-based end-effector control; However, for tasks like PickCube-v1, in which the robotic arm cannot easily remain static at the goal, we use the delta end-effector position control mode to achieve better performance. TD-MPC2 baseline also has both a sample-efficient set of environment configurations. The sample-efficient setting uses 1 CPU environment whereas the wall-time efficient setting uses 32 GPU parallelized environments. Less parallel environments relative to the number of updates per frame of data sampled generally lead to better sample efficiency.

\clearpage
\section{Imitation Learning Baselines}
\label{appendix:il_baselines}

\subsection{Behavior Cloning Baselines}

We evaluate three offline imitation learning baselines on PickCube, PushCube, StackCube, and PegInsertionSide using state and RGB observations. The baselines are Behavior Cloning (BC), Diffusion Policy (DP), and Action Chunking Transformer (ACT). Motion planning is used for the data sources. Table \ref{table:il_state_results} and \ref{table:il_rgb_results} show the best success rates obtained during training on different tasks for state and RGB observations, respectively. Overall we find that Diffusion Policy performs the best, especially when there are few demonstrations.

\begin{table*}[ht]
\centering
\setlength{\tabcolsep}{8pt}
\begin{tabular}{lllllcccc}
\toprule
 &  \multicolumn{2}{c}{\textbf{PickCube-v1}} & \multicolumn{2}{c}{\textbf{PushCube-v1}} & \multicolumn{2}{c}{\textbf{StackCube-v1}} & \multicolumn{2}{c}{\textbf{PegInsertionSide-v1}} \\ 
\cmidrule(lr){2-3}
\cmidrule(lr){4-5}
\cmidrule(lr){6-7}
\cmidrule(lr){8-9}
& Demos & Success & Demos & Success & Demos & Success & Demos & Success \\
\midrule
BC & 1000 & 0.03 & 1000 & 0.81 & 1000 & 0.00  & 1000 & 0.00 \\
\addlinespace
ACT & 1000 & 1.00 & 1000 & 0.97 & 1000 & 0.97 & 1000 & 0.43 \\
\addlinespace
DP  & 1000  & 1.00 & 1000 & 0.96 & 1000 & 0.99 & 1000 & 0.66 \\
\midrule
BC & 100 & 0.00 & 100 & 0.69 & 100 & 0.00  & 100 & 0.00 \\
\addlinespace
ACT & 100 & 0.70 & 100 & 0.99 & 100 & 0.50 & 100 & 0.14 \\
\addlinespace
DP  & 100  & 1.00 & 100 & 0.95 & 100 & 0.90 & 100 & 0.38 \\
\bottomrule
\end{tabular}
\caption{Comparison of imitation learning baselines on different tasks using state observations with varying number of demonstrations.}
\label{table:il_state_results}
\end{table*}

\begin{table*}[ht]
\centering
\setlength{\tabcolsep}{8pt}
\begin{tabular}{lllllcccc}
\toprule
 &  \multicolumn{2}{c}{\textbf{PickCube-v1}} & \multicolumn{2}{c}{\textbf{PushCube-v1}} & \multicolumn{2}{c}{\textbf{StackCube-v1}} & \multicolumn{2}{c}{\textbf{PegInsertionSide-v1}} \\ 
\cmidrule(lr){2-3}
\cmidrule(lr){4-5}
\cmidrule(lr){6-7}
\cmidrule(lr){8-9}
& Demos & Success & Demos & Success & Demos & Success & Demos & Success \\
\midrule
BC & 1000 & 0.03 & 1000 & 0.81 & 1000 & 0.00  & 1000 & 0.00 \\
\addlinespace
ACT       & 1000 & 0.98 & 1000 & 0.89 & 1000 & 0.80 & 1000 & 0.00 \\
\addlinespace
DP  & 1000  & 1.00 & 1000 & 0.86 & 1000 & 0.81 & 1000 & 0.00 \\
\midrule
BC & 100 & 0.00 & 100 & 0.00 & 100 & 0.00  & 100 & 0.00 \\
\addlinespace
ACT & 100 & 0.28 & 100 & 0.30 & 100 & 0.33 & 100 & 0.00 \\
\addlinespace
DP  & 100  & 0.76 & 100 & 0.41 & 100 & 0.61 & 100 & 0.00 \\
\bottomrule
\end{tabular}

\caption{Comparison of imitation learning baselines on different tasks using RGB observations with varying number of demonstrations.}
\label{table:il_rgb_results}
\end{table*}

\subsection{Vision-Language Action (VLA) Model Baselines}

We support training and/or evaluation of various vision-language action (VLA) model baselines. For evaluation, we support Octo and RT-x via the GPU parallelized version of the SIMPLER \cite{li24simpler} project. For RDT-1B, we support fine-tuning RDT-1B on a few ManiSkill3 demonstrations to then be evaluated on various tasks.

For both training and evaluation, a more classical sense-plan-act style method PerAct\cite{shridhar2022peract} is provided as a baseline in ManiSkill3. Different from RGB-based baselines like Octo and RT-x, PerAct operates on voxelized inputs. Our implementation follows the original PerAct design with modifications to demonstration generation and action execution: We obtain demonstrations by replaying trajectories with point cloud observations, which are voxelized directly during inference. This approach eliminates the need to explicitly convert RGBD observations to point clouds, as required in the original implementation. We hard-coded text descriptions for each task based on ManiSkill3 task cards. During inference, the end-effector pose-based actions are converted into joint position sequences using MPlib, which are then executed under ManiSkill3's joint-position (pd\_joint\_pos) control mode.

\begin{figure}[h]
    \centering
    \includegraphics[width=\linewidth]{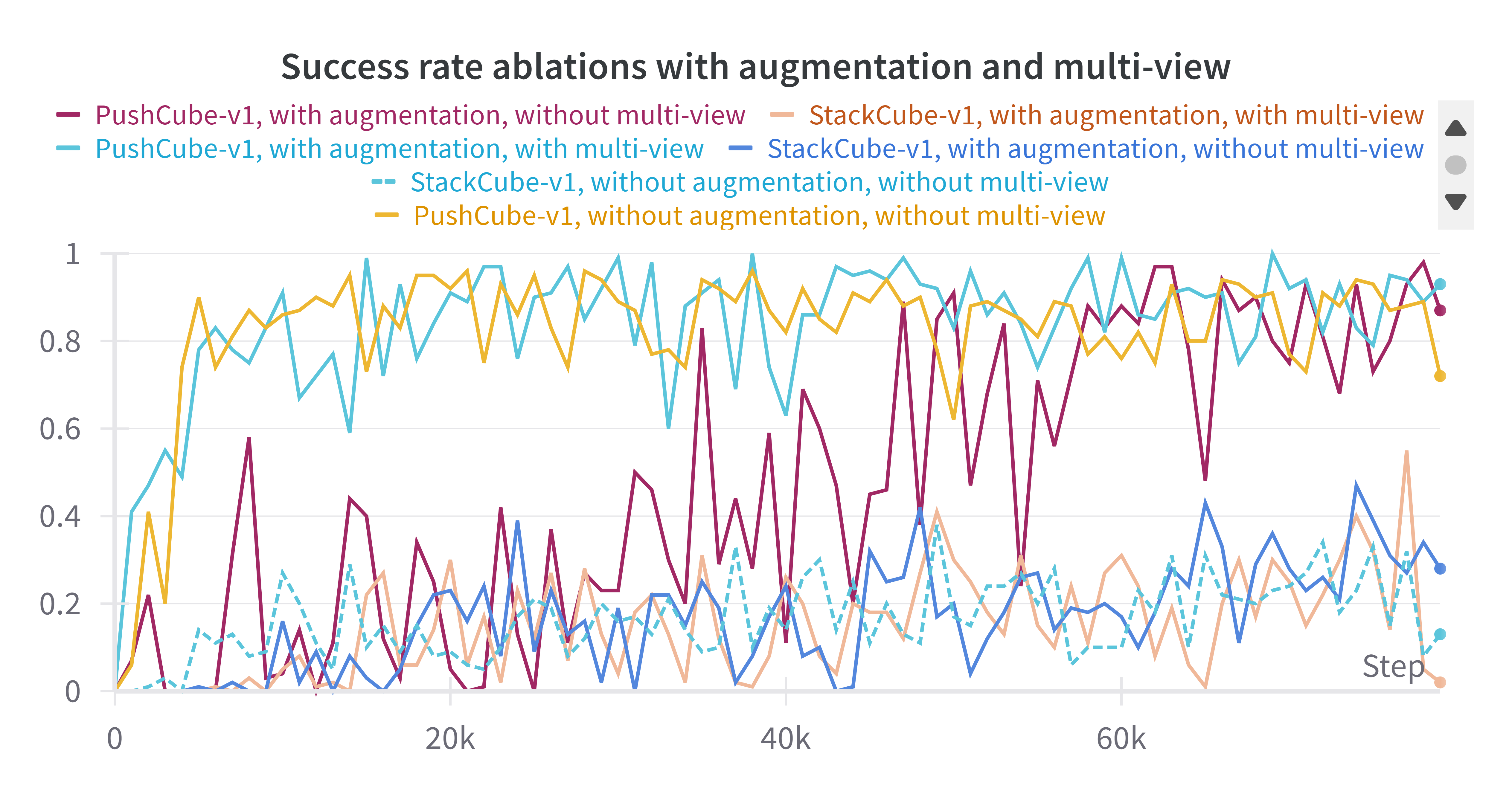}
    \caption{Success rate curves of PerAct over 80k training steps on PushCube-v1 and StackCube-v1}
    \label{fig:peract-success-rate-steps}
\end{figure}

We trained PerAct on PushCube-v1 and StackCube-v1 using 50 demonstration trajectories and evaluated over 100 episodes. The results are presented in Table \ref{tab:best-success-rates-reversed} and Figure \ref{fig:peract-success-rate-steps}.

\subsubsection{Multi-View and SE(3) Augmentation}

\begin{figure}[h]
    \centering
    \begin{subfigure}[b]{0.48\linewidth}
        \centering
        \includegraphics[width=\linewidth]{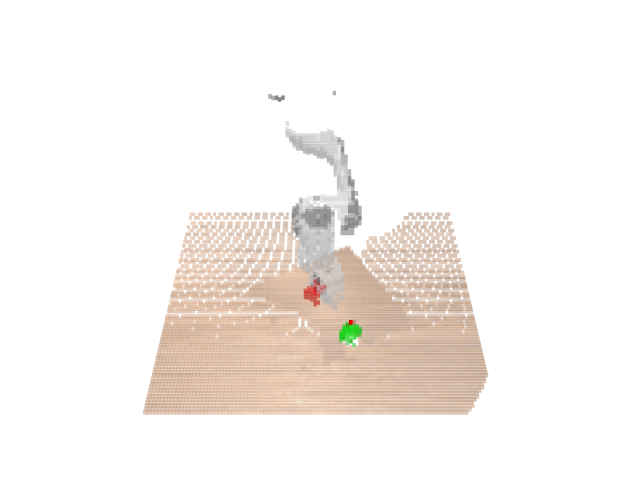}
        \caption{Default-view voxelized scene.}
        \label{fig:voxelized-single-view}
    \end{subfigure}
    \hfill
    \begin{subfigure}[b]{0.48\linewidth}
        \centering
        \includegraphics[width=\linewidth]{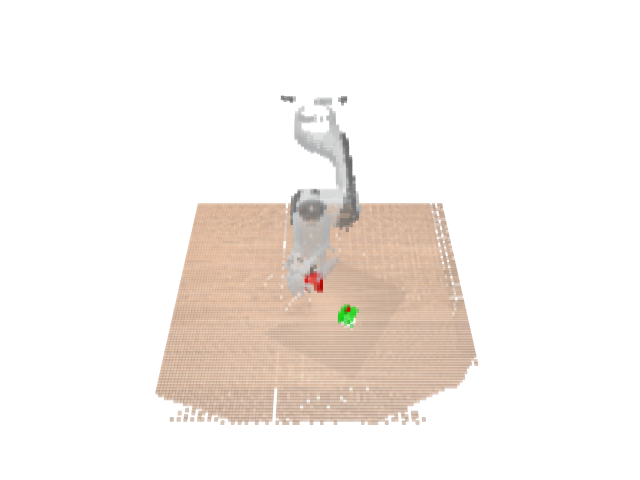}
        \caption{Multi-view voxelized scene.}
        \label{fig:voxelized-multi-view}
    \end{subfigure}
    \caption{Comparison between default-view and multi-view voxelized images in StackCube-v1. The red dot indicates the predicted action.}
    \label{fig:voxelized-comparison}
\end{figure}

\begin{figure}[h]
    \centering
    \begin{subfigure}[b]{0.4\linewidth}
        \centering
        \includegraphics[width=\linewidth]{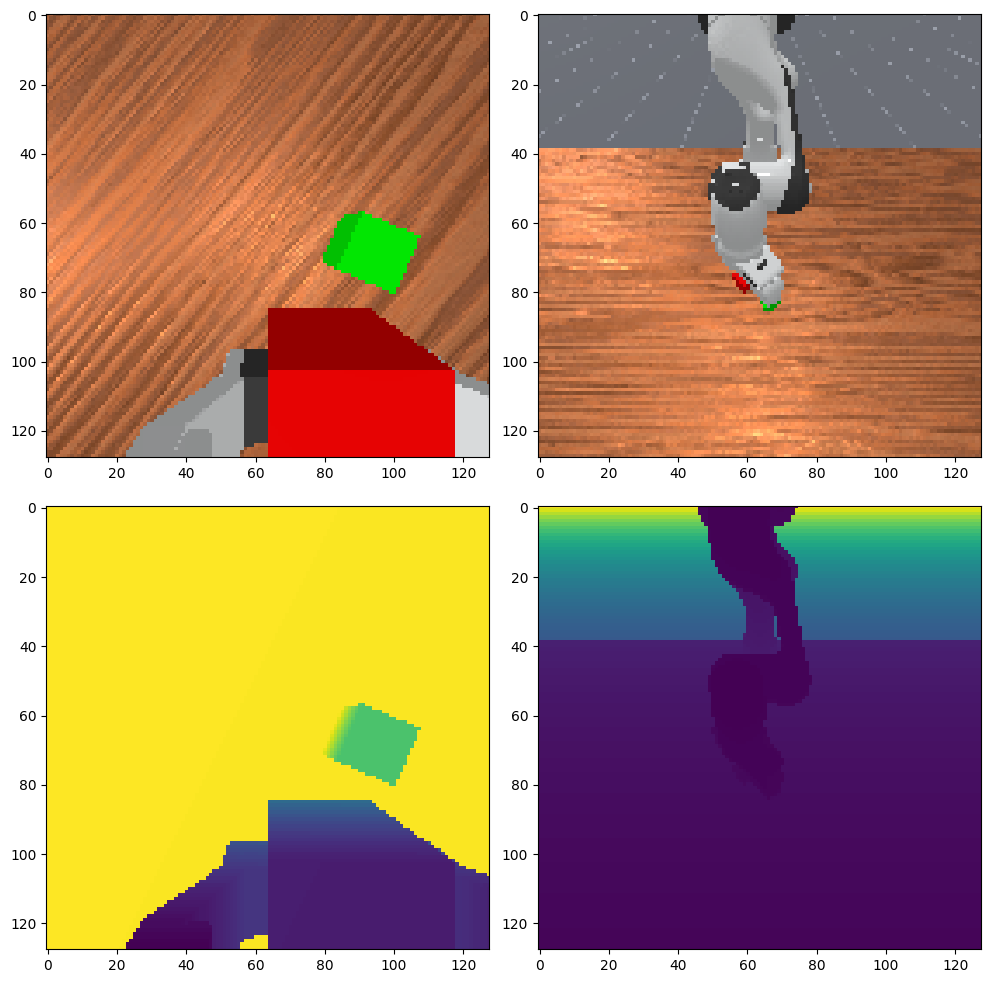}
        \caption{Default-view RGBD observations. Cameras from left to right: hand\_camera, base\_camera}
        \label{fig:rgbd-single-view}
    \end{subfigure}
    \hfill
    \begin{subfigure}[b]{0.55\linewidth}
        \centering
        \includegraphics[width=\linewidth]{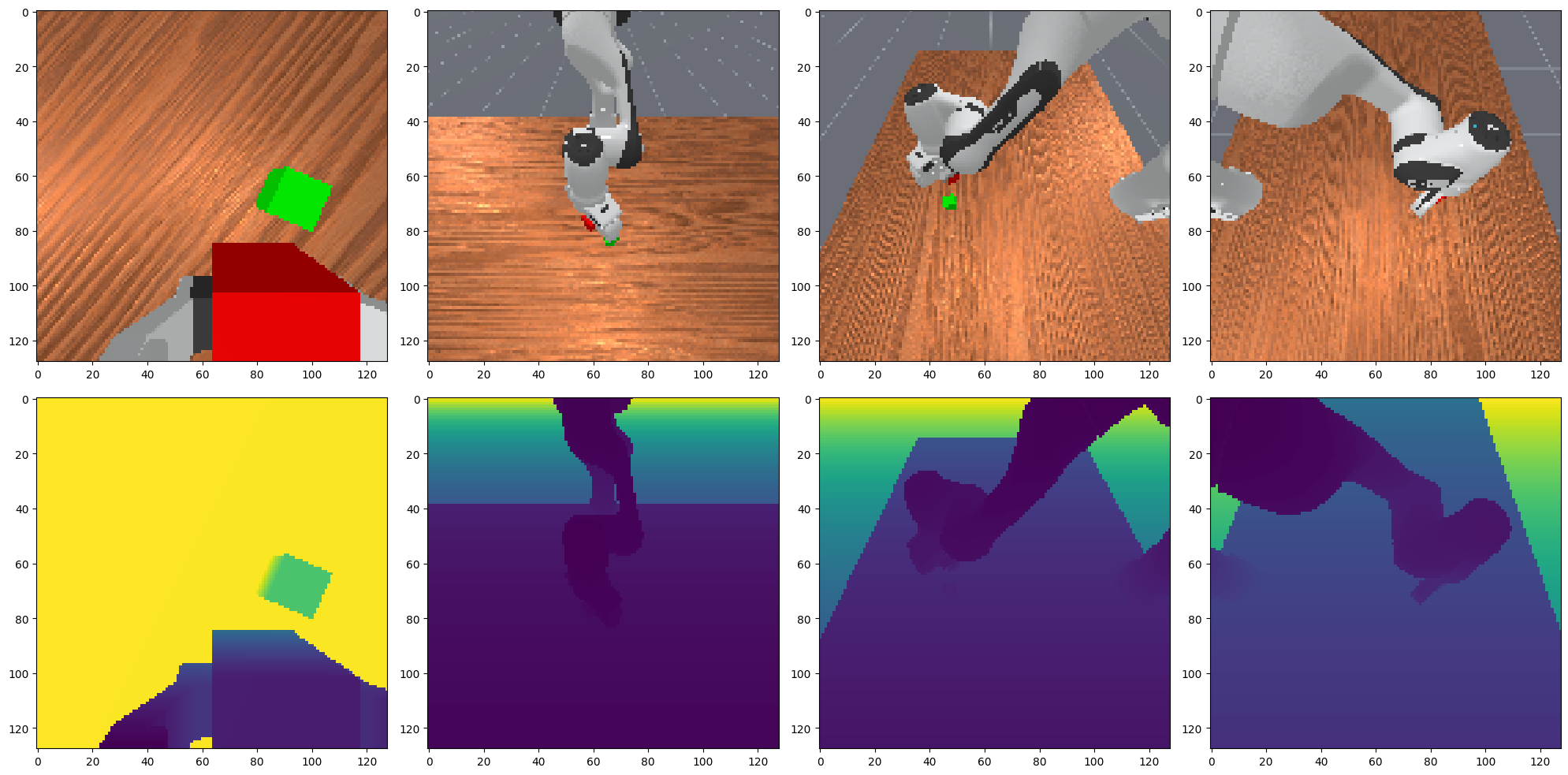}
        \caption{Multi-view RGBD observations. From left to right: hand\_camera, base\_camera, left\_shoulder\_camera, right\_shoulder\_camera}
        \label{fig:rgbd-multi-view}
    \end{subfigure}
    \caption{Comparison between default-view and multi-view RGBD observations in StackCube-v1. Each column shows one camera's observations.}
    \label{fig:rgbd-comparison}
\end{figure}

We also reported ablation results under \textbf{multi-view} and \textbf{SE(3) augmentation} settings: The \textbf{multi-view settings} include 4 cameras: a base camera, a left shoulder camera on the left of the arm, a right shoulder camera on the right, and a wrist camera mounted on the arm. Figures \ref{fig:voxelized-comparison} and \ref{fig:rgbd-comparison} illustrate that multi-view cameras provide finer voxelization details. Yet the difference for StackCube-v1 is marginal since the cubes and their surroundings are already visible via the hand and base cameras. Table \ref{tab:peract-cameras} details the cameras used for each experiment.

\begin{table}[h!]
\centering
\begin{tabular}{lcc}
\toprule
\textbf{Setting} & \textbf{PushCube-v1} & \textbf{StackCube-v1} \\
\midrule
No aug., no m.v. & 0.96 & 0.38 \\
aug., no m.v. & 0.98 & 0.47 \\
aug., m.v. & 1.00 & 0.55 \\
\bottomrule
\end{tabular}
\caption{Best success rates of PerAct over 80k training steps for different tasks and settings. aug.: demo augmentation. m.v.: multi-view}
\label{tab:best-success-rates-reversed}
\end{table}

\begin{table}[h!]
\centering
\begin{tabular}{lcc}
\toprule
\textbf{Task} & \textbf{Default Cameras} & \textbf{Multi-View Cameras} \\
\midrule
PushCube-v1 & base & base, l\_shldr, r\_shldr, wrist \\
StackCube-v1 & base, wrist & base, l\_shldr, r\_shldr, wrist \\
\bottomrule
\end{tabular}
\caption{Camera configurations for each task and setting. l\_shldr = left\_shoulder, r\_shldr = right\_shoulder.}
\label{tab:peract-cameras}
\end{table}

The \textbf{SE(3) augmentation} follows the PerAct approach, perturbing point clouds and actions with random translations ([$\pm$0.025m $\pm$0.025m $\pm$0.025m]) and rotations ([$\pm$0$^\circ$ $\pm$0$^\circ$ $\pm$5$^\circ$]).

To summarize the observations and results: (1) SE(3) augmentation helps improve the success rate, though making convergence slower. (2) Multi-view observations can speed up convergence and enhance success rates by providing a more comprehensive 3D scene understanding. (3) As the number of cameras increases, the point cloud size grows. However, voxelized inputs help maintain consistent GPU memory consumption despite the increased data size.

\clearpage

\section{Simulation and Rendering Benchmarking}
\label{appendix:sim-render-benchmark}

We carefully analyze the performance of ManiSkill3's parallel simulation and simulation+rendering performance compared to Isaac Lab. Currently we only have accurate results for a simple cartpole environment. While one can try and compare more complex environments it is difficult to align things perfectly. We compare against the Isaac Lab v1.2.0 which was first version with parallel rendering support that was released a few months after we initially released our framework.

\subsection{Setup}

To keep things as fair as possible as both Isaac Lab and ManiSkill3 use PhysX, we ensure the following simulation configurations are the same:
\begin{itemize}
    \item Simulation frequency: 120
    \item Control frequency: 60
    \item Solver Position Iterations: 4
    \item Solver Velocity Iterations: 0
\end{itemize}

We further try to ensure the objects in the scene are as similar as possible. Finally, reward/termination computations are explicitly removed to not factor in benchmark timings. Tests were conducted on a RTX 4090 GPU.
\begin{figure}[h]
    \centering
    \includegraphics[width=\linewidth]{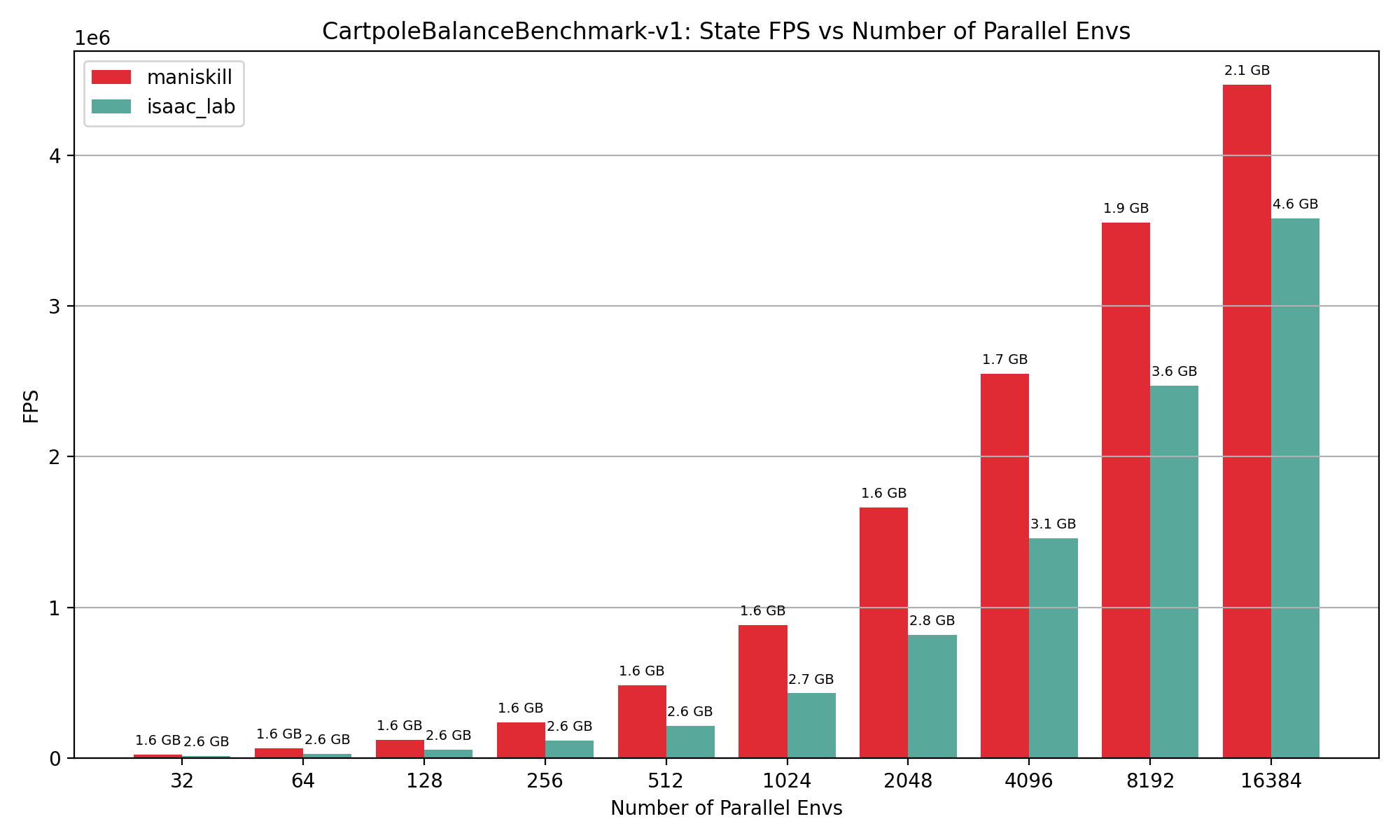}
    \caption{Simulation FPS against number of parallel environments without rendering. Annotated numbers on top of data points indicate the GPU memory usage. ManiSkill speed for this task is on par with Isaac Lab.}
    \label{fig:parallel_envs_ablation}
\end{figure}

\subsection{Simulation Only Benchmark Results}

Testing on the cart pole task, we see that ManiSkill3 runs a little faster and uses about 1.5x to 2x less GPU memory in Figure \ref{fig:parallel_envs_ablation}. We note that while ManiSkill3 appears faster here, one could make many simulation specific optimizations (simplified collisions meshes, tuned physics solver configurations etc.) to increase simulation speed and trade off simulation accuracy. We further note it is difficult to make truly apples-to-apples comparisons between simulators and that this is simply just one data point and may not extend to other environments. We specifically create documentation/tutorials on the various tricks/optimizations a user can perform to improve simulation speed and/or fidelity from the environment object/robot models to simulation configurations.

\subsection{Simulation+Rendering Benchmark Results}

We ablate on a number of aspects of visual data collection where we simulate and render an environment. We record the FPS of taking 1000 random actions in the environment and fetching the 1000 visual observations. Overall Isaac Lab shows about 2-4x higher GPU memory usage compared to ManiSkill3, with the minimum amount of GPU memory taken up by just environments with cameras enabled being 4.8GB compared to 1.7GB in ManiSkill3.

\subsubsection{Ablation on Realistic Camera Settings}

\begin{figure}[h]
    \centering
    \includegraphics[width=\linewidth]{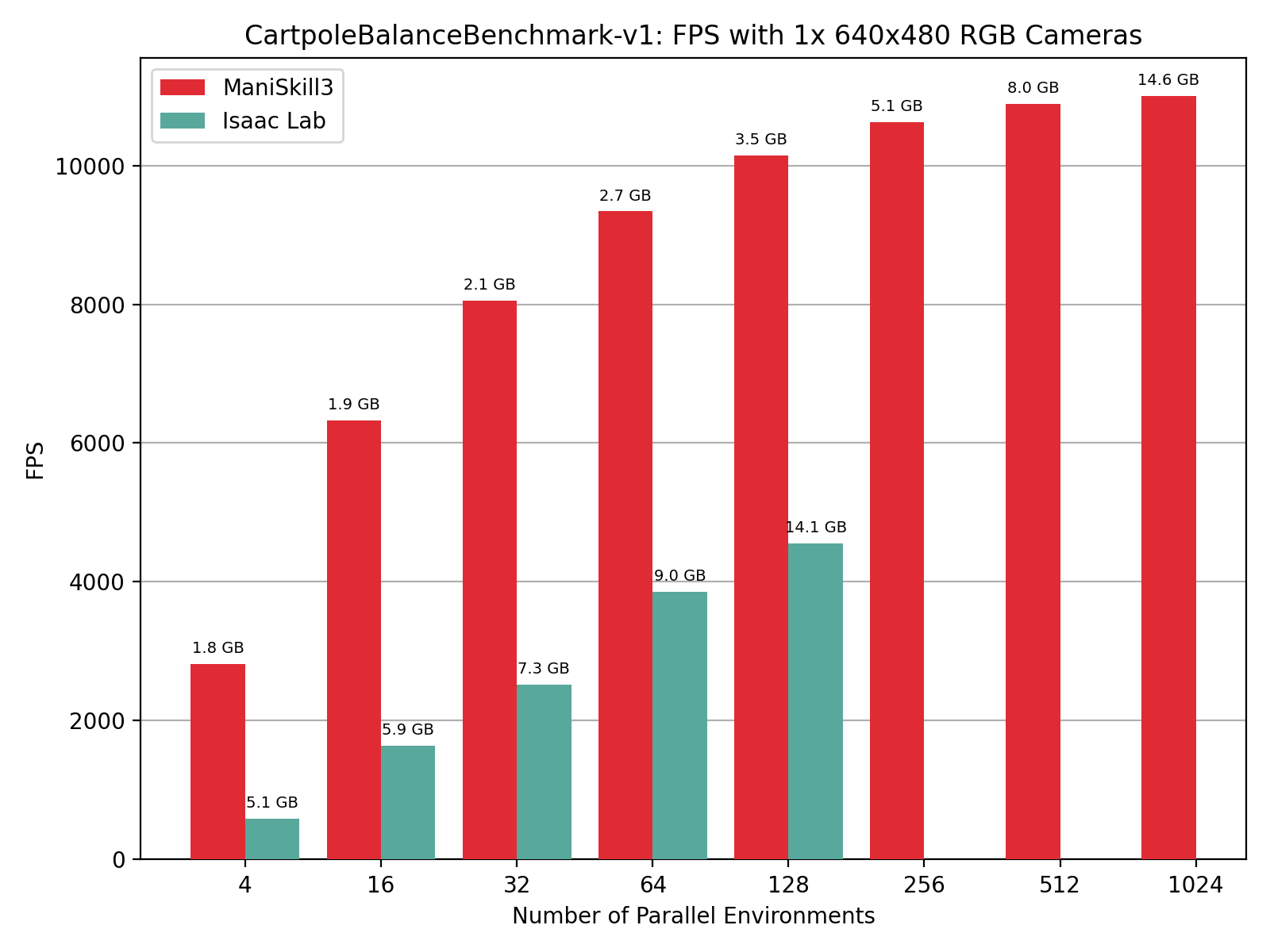}
    \caption{Simulation+Rendering of RGB or Depth FPS against number of parallel environments with 1x640x480 camera. Annotated numbers on top of data points indicate the GPU memory usage.}
    \label{fig:rt_dataset_rendering_performance}
\end{figure}

\begin{figure}[h]
    \centering
    \includegraphics[width=\linewidth]{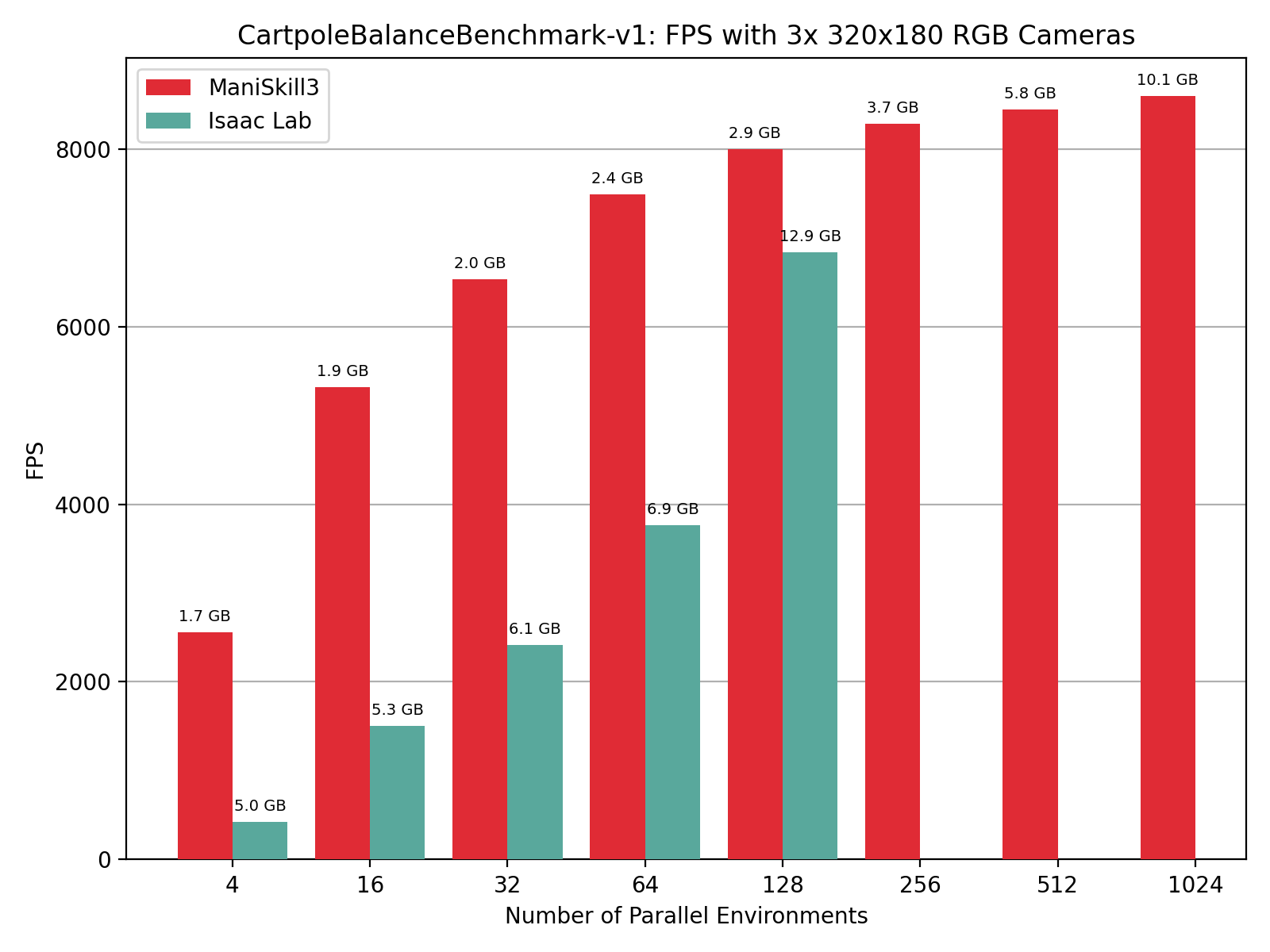}
    \caption{Simulation+Rendering of RGB or Depth FPS against number of parallel environments with 3x320x180 cameras. Annotated numbers on top of data points indicate the GPU memory usage.}
    \label{fig:droid_dataset_rendering_performance}
\end{figure}

We test on two realistic camera resolutions/setups based on the setup of the Open-X dataset \citep{open_x_embodiment_rt_x_2023} and the Droid dataset \citep{khazatsky2024droid}. Many datasets in Open-X like the Google RT datasets have a single 640x480 resolution RGB camera, and the Droid dataset uses three 320x180 resolution RGB cameras. The results are shown in Figures \ref{fig:rt_dataset_rendering_performance} and \ref{fig:droid_dataset_rendering_performance}. In both cases ManiSkill3 has about 2-4x better GPU memory efficiency and runs about 2x faster. Note that Isaac Lab's rendering output is a bit different from ManiSkill3, we share some qualitative examples in Appendix \ref{appendix:qualitative-comparison-rendering}. We further note that both ManiSkill3 and Isaac Lab support more photorealistic, high quality renders with full ray-tracing that is not parallelized. Example ManiSkill3 high-quality ray-traced renders can be seen in Figure \ref{fig:task_categories}.

\subsubsection{Ablation on Camera Size}

\begin{figure*}[!h]
    \centering
    \includegraphics[width=0.8\linewidth]{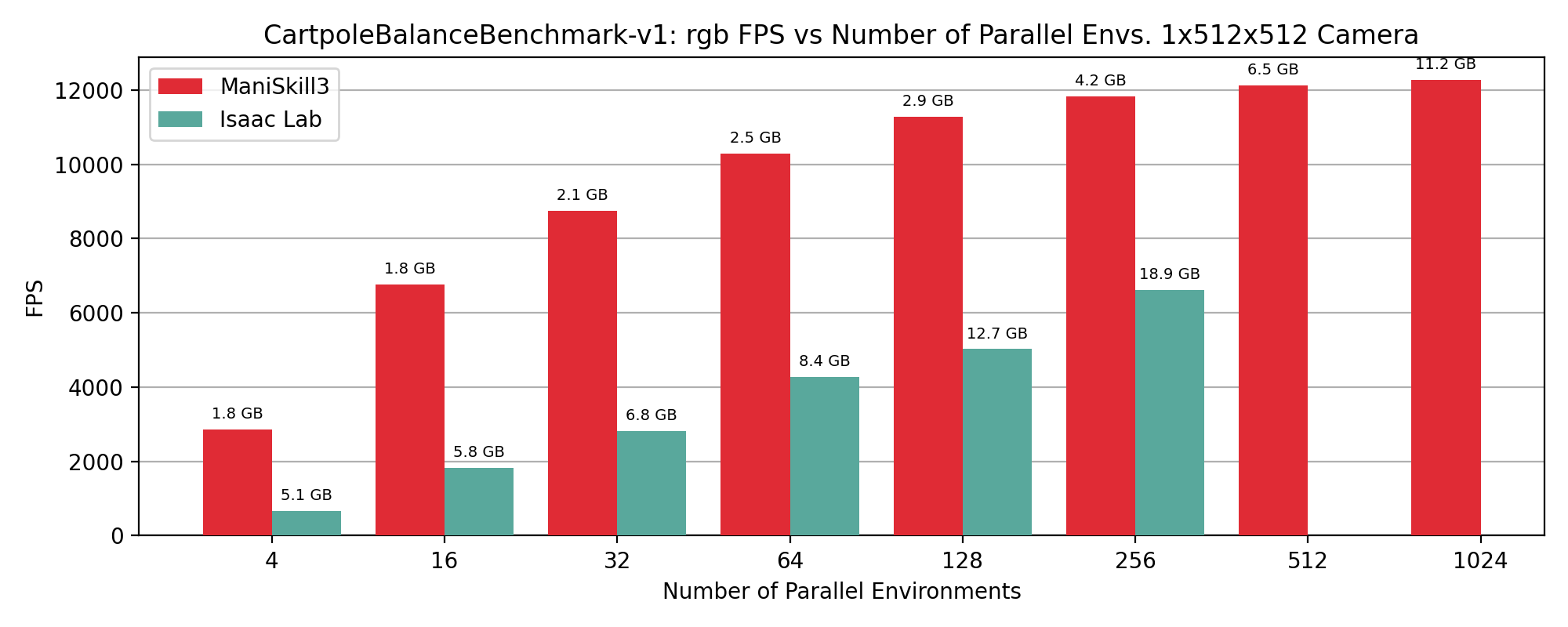}
    \caption{Simulation+Rendering (RGB) FPS against number of parallel environments 1x512x512 camera. Annotated numbers on top of data points indicate the GPU memory usage.}
    \label{fig:512x512perf}
\end{figure*}

\begin{figure*}[!h]
    \centering
    \includegraphics[width=0.8\linewidth]{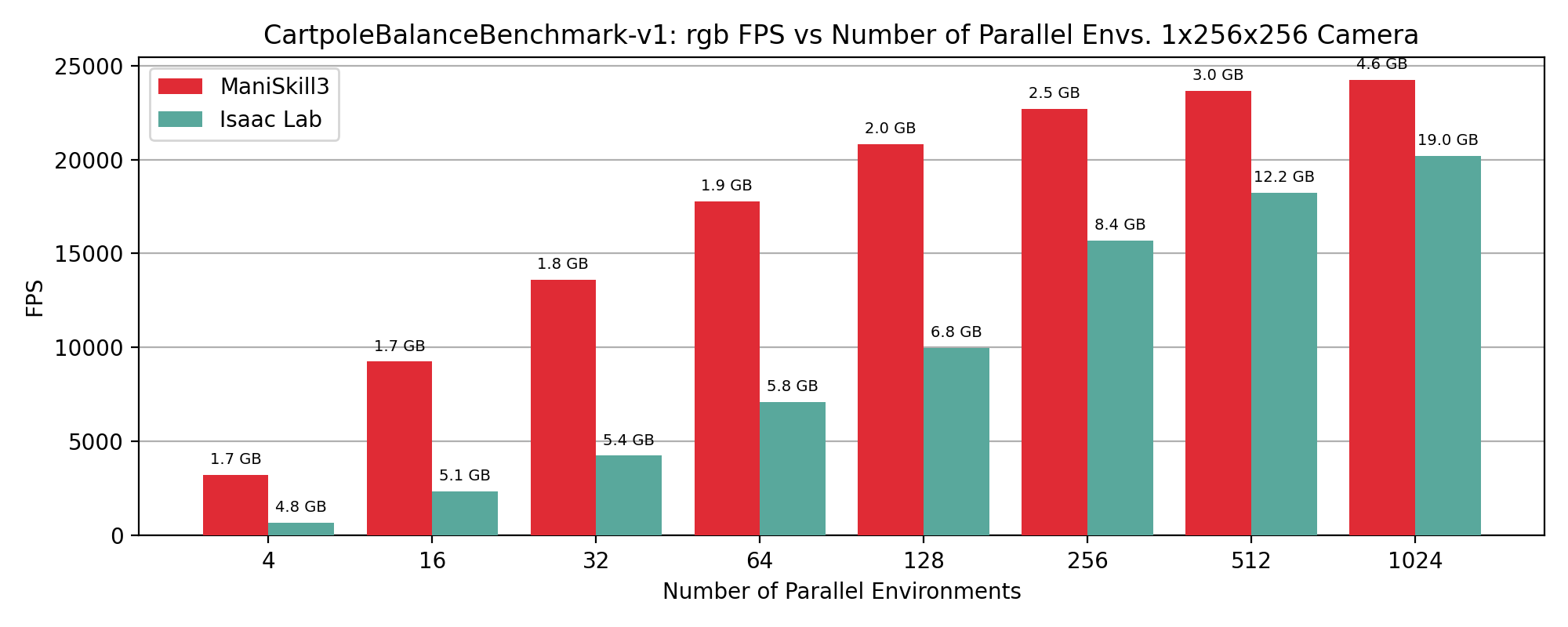}
    \caption{Simulation+Rendering (RGB) FPS against number of parallel environments 1x256x256 camera. Annotated numbers on top of data points indicate the GPU memory usage.}
    \label{fig:256x256perf}
\end{figure*}

\begin{figure*}[!h]
    \centering
    \includegraphics[width=0.8\linewidth]{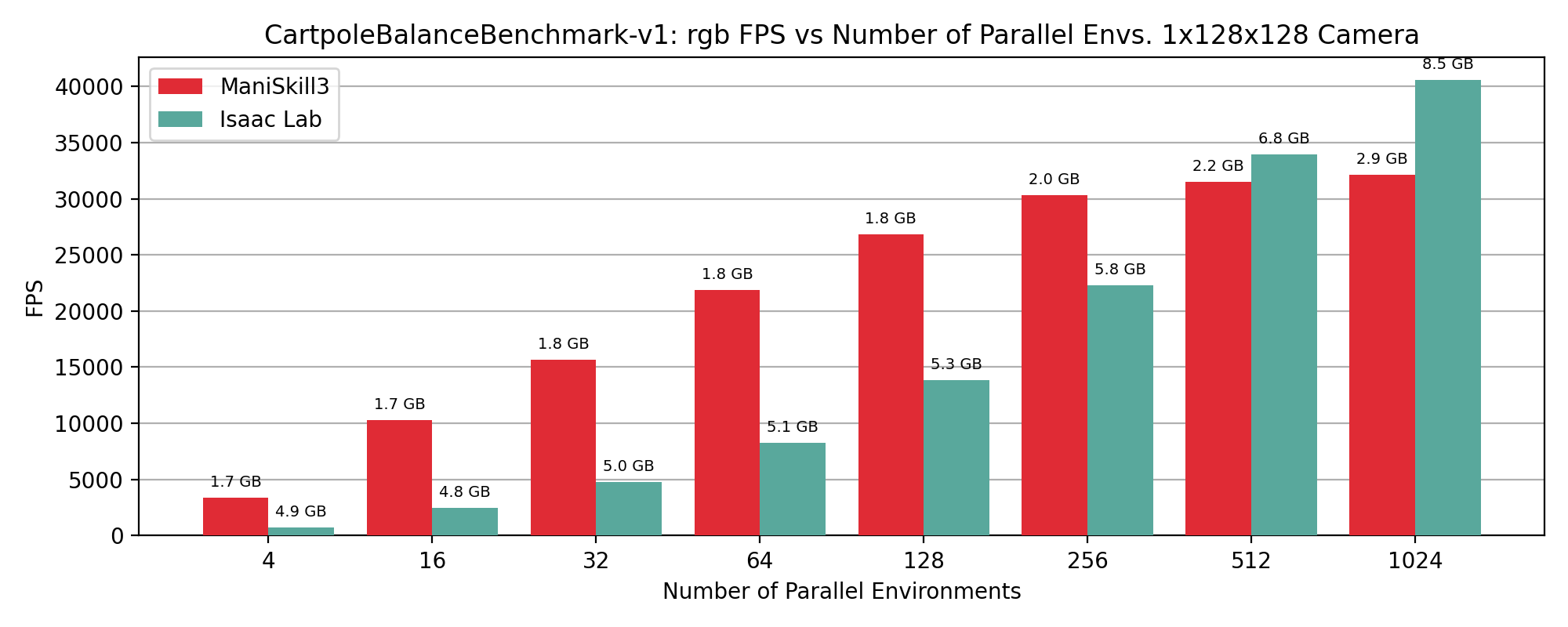}
    \caption{Simulation+Rendering (RGB) FPS against number of parallel environments 1x128x128 camera. Annotated numbers on top of data points indicate the GPU memory usage.}
    \label{fig:128x128perf}
\end{figure*}

We ablate on square camera sizes from a large 512x512 resolution to a small 128x128 resolution shown in Figures \ref{fig:512x512perf} \ref{fig:256x256perf} \ref{fig:128x128perf}. We observe that Isaac Lab runs at most about 1.25x faster when there are a high number of parallel environments with small camera resolutions compared to ManiSkill3. At larger camera resolutions ManiSkill3 outperforms up to 2x in speed and 4x in GPU memory usage. Notably, ManiSkill3 always outperforms about 2-4x in speed and 2-3x in GPU memory usage for smaller number of parallel environments. While Isaac Lab is fast at smaller resolutions, we note that some manipulations tasks are fairly impractical and impossible to solve at small resolutions. Moreover memory efficiency is absolutely critical for RL applications that can remain fast by keeping large replay buffers on the GPU instead of using CPU memory. For this reason all visual RL baselines in ManiSkill3 typically do not use more than 256 to 1024 parallel environments as RL rollouts only run marginally faster with more environments beyond that point (for both Isaac Lab and ManiSkill3) but GPU memory use worsens a lot more.

\subsection{Qualitative Rendering Results}
\label{appendix:qualitative-comparison-rendering}

\begin{figure*}[h]
    \centering
    \includegraphics[width=\linewidth]{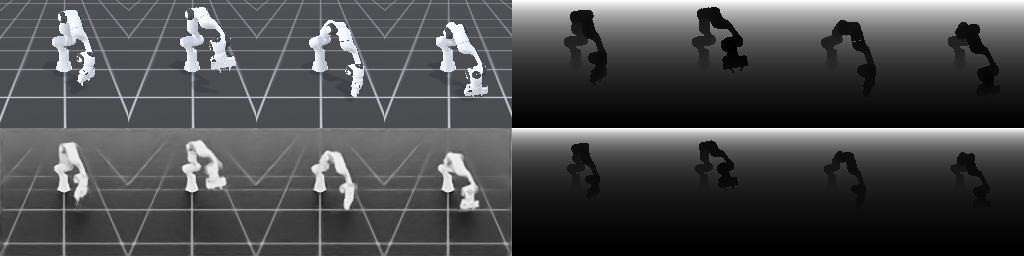}
    \caption{Comparison of ManiSkill3 (Top row) and Isaac Lab (Bottom Row) parallel rendering 128x128 RGB+Depth image outputs of the Franka Panda arm.}
    \label{fig:franka-visual-comparison_small}
\end{figure*}

\begin{figure*}[h]
    \centering
    \includegraphics[width=\linewidth]{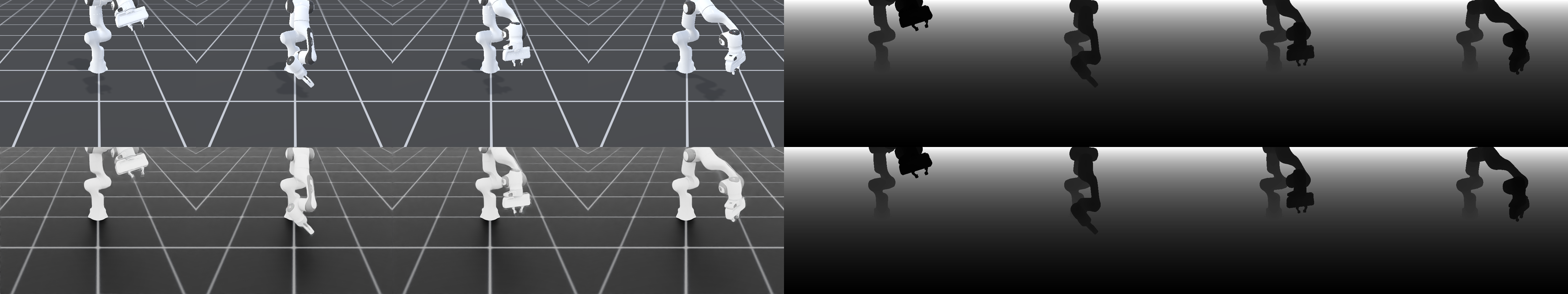}
    \caption{Comparison of ManiSkill3 (Top row) and Isaac Lab (Bottom row) parallel rendering 640x480 RGB+Depth image outputs of the Franka Panda arm.}
    \label{fig:franka-visual-comparison_large}
\end{figure*}

To the best of our capabilities we try to make the benchmarked environments look as similar as possible in terms of visuals. It is not possible to keep everything the same given fundamental differences in the parallel rendering system of Isaac Lab compared to ManiSkill3 or easily quantify the rendering differences. Section \ref{sec:sim+rendering} details a little bit about the differences. Figure \ref{fig:cartpole-visual-comparison} in the main paper shows the cartpole RGB+Depth rendering results. Figures \ref{fig:franka-visual-comparison_small} and \ref{fig:franka-visual-comparison_large} shows RGB renders of a environment with a Franka Panda arm with different resolutions.

\clearpage
\section{VR Teleoperation}
\label{appendix:teleop}
ManiSkill3 provides VR support for all mainstream VR devices by implementing the OpenVR client protocol designed by Steam. Specifically, ManiSkill3 receives camera intrinsic and extrinsic parameters of the head-mounted display, VR controllers' poses, and the operator's hand and wrist poses if supported by the hardware; and it sends stereo video streams at 4K resolution for OpenVR to display in the headset. Under the hood, OpenVR communicates with SteamVR and ALVR which translate hardware-dependent VR implementation into the unified OpenVR client protocol.

The VR feedback loop runs at 60 Hz which is crucial for smooth user experience, while an asynchronous action translator translates sensed VR poses to robot actions at 20 Hz, so we can tolerate computationally intensive translation algorithms. 

\begin{figure}[hp]
    \centering
    \includegraphics[width=0.9\linewidth]{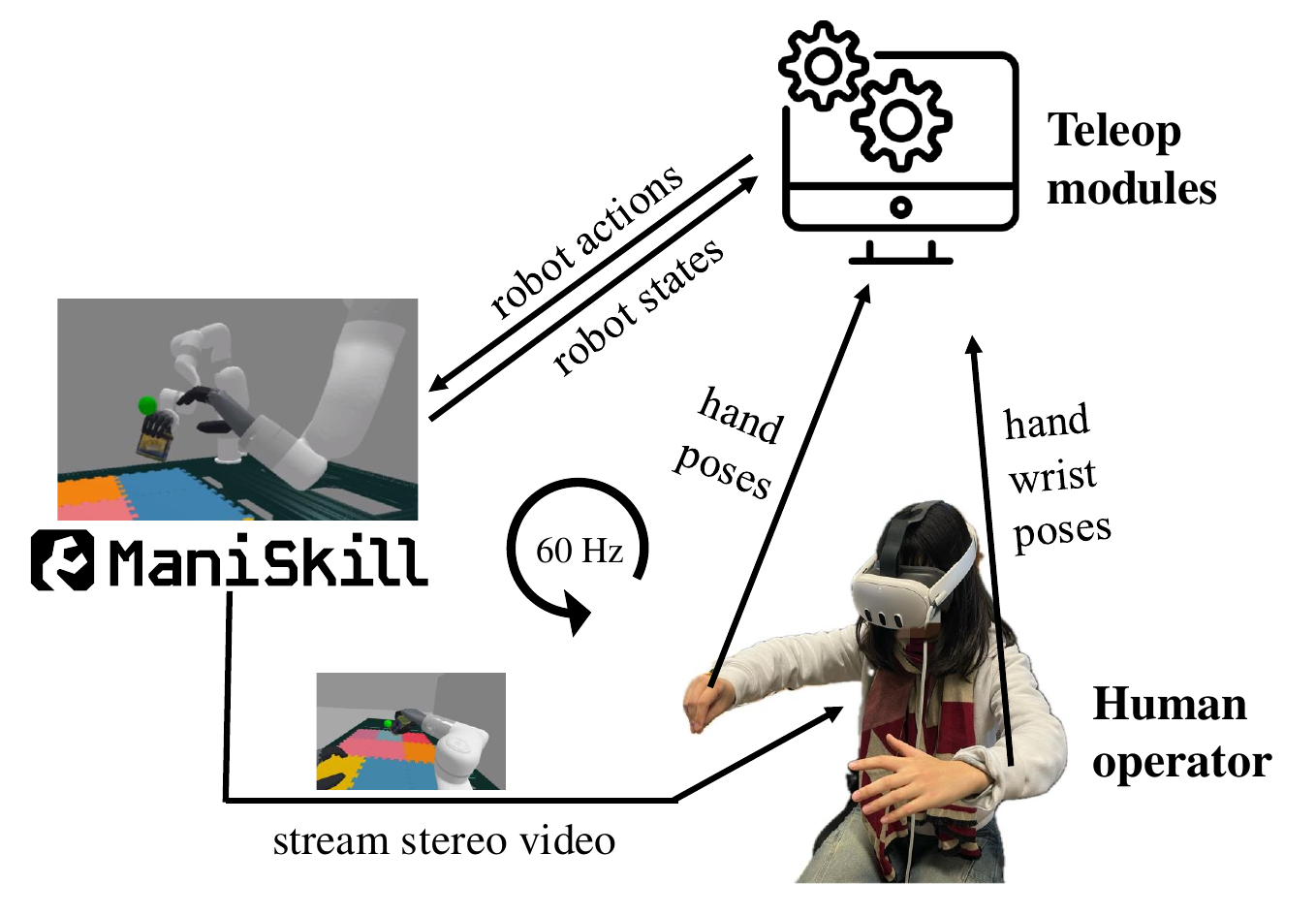}
    \caption{Overview of our VR teleoperation system. The operator controls the robot's arm and hand through real-time tracking of wrist and hand poses, while receiving stereo video feedback. The VR devices stream human pose data to a server, which retargets and sends joint commands to the robot. }
    \label{fig:vr_overview}
\end{figure}

Overview for our VR teleoperation system is shown in Figure \ref{fig:vr_overview}.  In a simulation setup, the stereo video is rendered based on simulated environments. For real-world robot teleoperation, the stereo video is generated from point clouds captured by depth cameras. We utilize the SAPIEN engine to achieve high-speed rendering of these high-resolution stereo videos. To simplify the complex SteamVR setup, we provide a Docker image for a smoother and faster setup process for users.

\begin{figure*}[tp]
    \centering
    \includegraphics[width=0.9\linewidth]{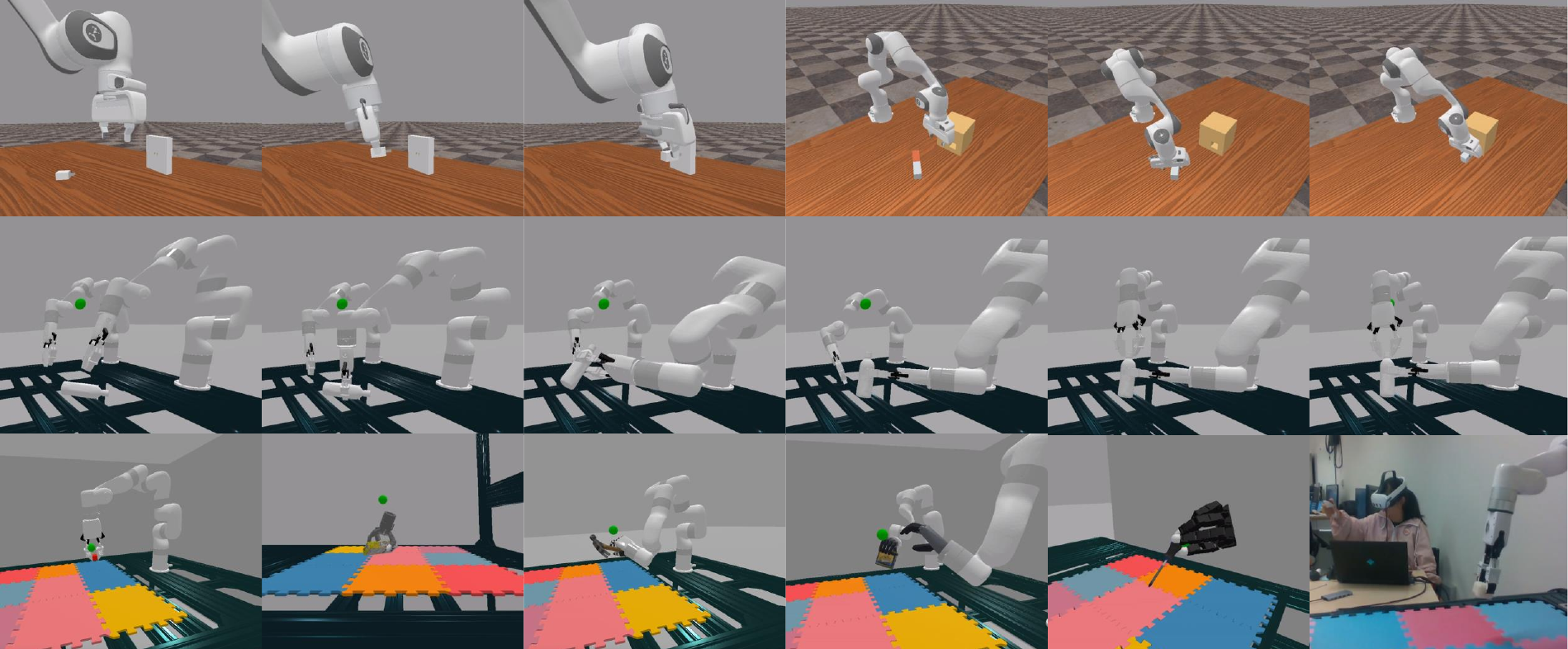}
    \caption{Demonstration of VR teleoperation across various tasks and robotic setups. Top row: Sequential images of a Panda arm performing precise insertions – charging connector insertion (left) and peg insertion (right) – showcasing accurate teleoperation for tasks requiring high precision. Middle row: Sequence of dual XArm7 robots with grippers demonstrating the grasping and unscrewing of a bottle, highlighting multi-arm coordinated teleoperation. Bottom row: Various teleoperation setups and manipulations, including (from left to right) XArm7 with a gripper, a floating Delto 3-finger hand, XArm7 with Inspire hand, dual XArm7 with Ability hand, a floating Allegro hand, and real-time teleoperation of an XArm7 with a gripper grasping objects.}
    \label{fig:vr-demo}
\end{figure*}

\subsection{Control}
\begin{figure}[htp]
\centering
\includegraphics[width=0.7\linewidth]{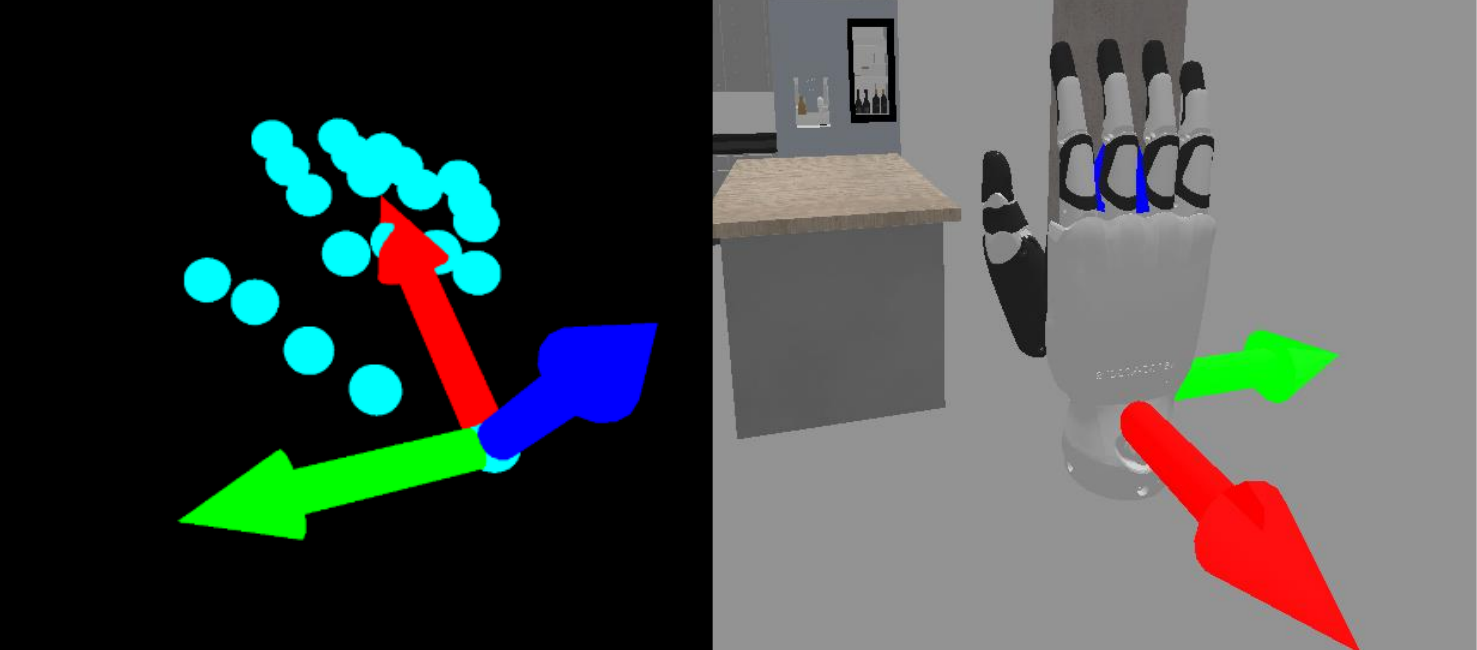}
\caption{Illustration of frame mismatch between human wrist poses and the robot's end-effector frame. Left: human wrist frame, Right: Inspire hand frame as an example.}
\label{fig:vr_ee_frame}
\end{figure}

The human-motion-to-robot-action conversion system consists of three main modules:

1) \textbf{Arm Control Module}, which converts human wrist poses into robot arm joint positions. However, directly mapping the absolute orientation and position of wrist poses to robot's end-effector poses can lead to strange behaviors. For example there can be a mismatch between the coordinate frames of human wrist poses and the robot's end-effector frame as illustrated in Figure \ref{fig:vr_ee_frame}. To address this, we provide configurations out-of-the-box for the most common robotics arms. Additionally, we offer a GUI and tools to assist users in quickly computing the transformation matrix for their customized robots. 
    
    Our system utilizes a modified version of the Closed-loop Inverse Kinematics (CLIK) algorithm, implemented with the Pinocchio library \citep{pinocchioweb, carpentier2019pinocchio}, to calculate the joint angles of the robot's arm. 
    
    The approach uniquely addresses the problem of inverse kinematics (IK) for two coupled end-effectors simultaneously, which is critical in dual-arm robots with a sliding joint at the shoulders. 
    
    For a robot with $n$ joints and two coupled end-effectors $EE_i$, we aim to find joint position $q$ that minimizes the pose error for both end-effectors relative to their respective target poses $T_i$. Solving IK independently for two end-effectors suffers from large errors. Alternatively, iterative approaches usually result in slow convergence. Our solution extends the standard CLIK by computing a concatenated Jacobian that accounts for both end-effectors simultaneously. This enables faster and more accurate optimization of dual-arm configurations.

    Additionally, the system allows users to control the movement of specific joints during coordinated motion. Instead of using a binary mask (0 for no movement, 1 for full movement) to constrain joints, we employ a soft mask with values between 0 and 1. A lower mask value reduces motion in the corresponding joint, leading to a solution that minimizes undesired movement. To ensure smooth arm motions, we apply an SE(3) group filter to the input end-effector poses before the IK computations.

2) \textbf{Hand Control Module}, which translates human finger poses into corresponding robot hand joint positions. Following \citep{qin2023anyteleop, cheng2024open, ding2024bunny}, we formulate the hand motion retargeting process as an optimization problem. The objective function for this optimization is defined as follows:
    \begin{equation}
        \min_{q_t} \sum_{i=0}^N \Vert \alpha^i v_{t}^i - f_i(q_t)\Vert^2 + \beta \Vert q_t - q_{t-1} \Vert, 
    \end{equation}
    where $q_t$ denotes the robot hand joint positions at time $t$, $v_t^i$ is the $i$-th keypoint vector of human hand, and $f_i(q_t)$ gives the corresponding $i$-th keypoint vector of robot hand using forward kinematics with joint positions $q_t$. The scaling factor $\alpha_i$ compensates for the differences in hand size between the human and robot hands and treat each $i$-th keypoint differently as thumb finger size and pinky finger size can vary a lot, $\beta$ weights the regularization term to ensure temporal consistency between consecutive joint positions. The optimization is implemented by NLopt solver \cite{qin2023anyteleop}. 

    For dexterous robot hands, we map vectors from human hand fingertips to palm base to corresponding vectors on the robot hand. In certain cases, such as with the Inspire hand, additional vector mappings are used to improve motion accuracy--for example, an extra vector from the thumb metacarpophalangeal joint to the thumb tip is employed. For grippers, we use a single vector optimization between the human thumb and index fingertips, which corresponds to the gripper's upper and lower ends. This allows intuitive control over the gripper's opening and closing motions through simple pinching gestures with the operator’s index finger and thumb \cite{cheng2024open}.

    We fine-tune configurations for several common robots and provide calibration tools to assist users in adapting the system to their custom robots.
    
3) \textbf{Controller Control Module}, which is also used to control gripper motion, allowing simple and effective control of grippers. Though Apple Vision Pro does not include a physical controller, our system still supports controller-based input. By clipping the controller, users can trigger the closing action of the gripper, enabling intuitive and responsive control during operation.

\subsection{Sim-to-Real Interface}

Our system employs a unified interface for both simulation and real-world setups by aligning the robot's end-effector with the absolute position of human hand. This alignment helps teleoperators better understand spatial information. Previous approaches \cite{qin2023anyteleop, cheng2024open, ding2024bunny} struggled to achieve spatial alignment between the human hand and the robot hand, forcing operators to compensate for parallax effects, which made the teleoperation experience less intuitive.

A key challenge arises when attempting to align the human hand with the robot hand inside a VR headset, especially since the real robot may be spatially displaced in the physical world. To solve this, we project the point clouds captured by depth cameras positioned around the real robot into the VR headset. The camera poses are calibrated using EasyHec \cite{chen2023easyhec}. 

\begin{figure*}[htp]
    \centering
    \includegraphics[width=0.9\linewidth]{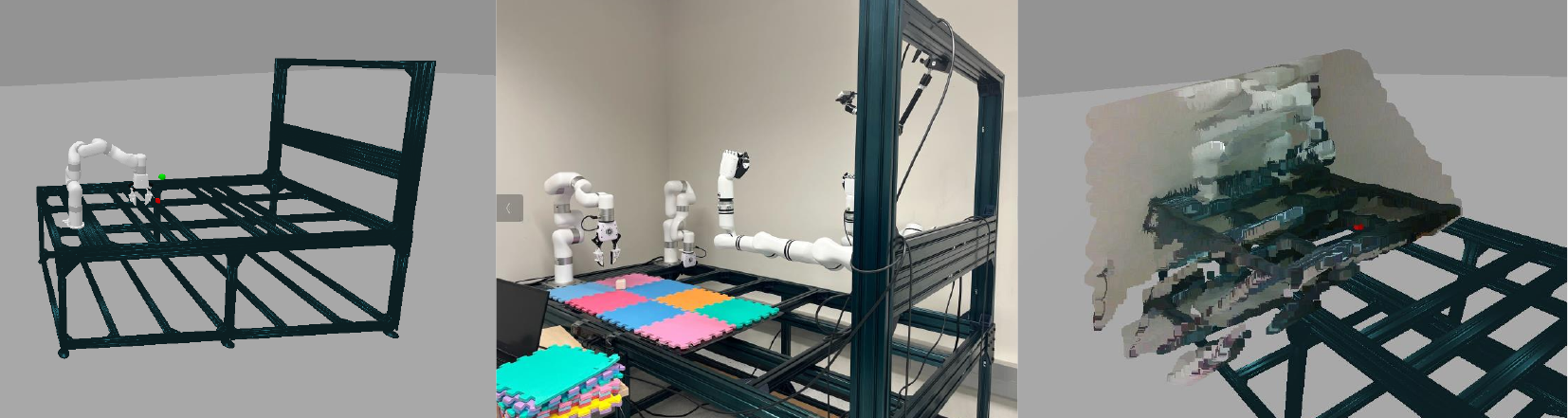}
    \caption{Illustration of spatial alignment of simulation and real-world environments. Left: simulation environment, where robot control are tested, Middle: real-world teleoperation setup, Right: point cloud captured from the depth cameras accurately aligned with the virtual robot in the simulation, demonstrating a "digital twin" setup.}
    \label{fig:vr_sim2real}
\end{figure*}

This setup ensures that both the simulation environment and the real-world point cloud are aligned in a "digital twin" manner as illustrated in Figure \ref{fig:vr_sim2real}. Though we do not require the visual textures to match the real world, critical elements such as the robot's position, forward and inverse kinematics, and control interface must be aligned. This alignment allows the same human control signals to produce identical robot actions in both the simulator and the real-world environment. Consequently, teleoperation in the real world becomes as intuitive and consistent as it is in simulation.

\end{document}